\documentclass[journal]{IEEEtran}

\usepackage[linesnumbered,ruled]{algorithm2e}
\usepackage{graphicx,amssymb,mathrsfs,amsmath,array,color}
\usepackage{setspace}
\usepackage{subfigure}
\usepackage{multirow}
\usepackage{booktabs}
\usepackage[colorlinks,urlcolor=blue,driverfallback=dvipdfm]{hyperref}
\usepackage{algpseudocode}

\newcommand{\norm}[1]{\lVert#1\rVert}

\makeatletter

\DeclareMathOperator{\tr}{tr}
\DeclareMathOperator{\F}{F}
\DeclareMathOperator{\T}{T}

\begin{document}
\title{Joint and Progressive Subspace Analysis (JPSA) with Spatial-Spectral Manifold Alignment for Semi-Supervised Hyperspectral Dimensionality Reduction}

\author{Danfeng Hong,~\IEEEmembership{Member,~IEEE,}
        Naoto Yokoya,~\IEEEmembership{Member,~IEEE,}
        Jocelyn Chanussot,~\IEEEmembership{Fellow,~IEEE,}
        Jian Xu,~\IEEEmembership{Member,~IEEE,}
        and~Xiao Xiang Zhu,~\IEEEmembership{Senior Member,~IEEE}
\thanks{This paper is an extended version in algorithm of \cite{hong2018joint} published in ECCV2018. The work of D. Hong is supported by the German Research Foundation (DFG) under grant ZH 498/7-2, and by the Helmholtz Association through the Framework of Helmholtz Artificial Intelligence Cooperation Unit (HAICU) - Local Unit ``Munich Unit @Aeronautics, Space and Transport (MASTr)''. The work of X. Zhu is jointly supported by the European Research Council (ERC) under the European Union's Horizon 2020 research and innovation programme (grant agreement No. [ERC-2016-StG-714087], Acronym: \textit{So2Sat}), by the Helmholtz Association through the Framework of Helmholtz Artificial Intelligence Cooperation Unit (HAICU) - Local Unit ``Munich Unit @Aeronautics, Space and Transport (MASTr)'' and Helmholtz Excellent Professorship ``Data Science in Earth Observation - Big Data Fusion for Urban Research'' and by the German Federal Ministry of Education and Research (BMBF) in the framework of the international future AI lab ``AI4EO -- Artificial Intelligence for Earth Observation: Reasoning, Uncertainties, Ethics and Beyond''. The work of N. Yokoya is supported by the Japan Society for the Promotion of Science (KAKENHI 18K18067). The work of J. Chanussot is supported by the AXA Research Fund. (\emph{Corresponding author: Xiao Xiang Zhu}).}
\thanks{D. Hong is with the Remote Sensing Technology Institute (IMF), German Aerospace Center (DLR), 82234 Wessling, Germany, and also with the Univ. Grenoble Alpes, CNRS, Grenoble INP, GIPSA-lab, 38000 Grenoble, France. (e-mail: hongdanfeng1989@gmail.com)}
\thanks{N. Yokoya is with Graduate School of Frontier Sciences, the University of Tokyo, 277-8561 Chiba, Japan, and also with the Geoinformatics Unit, RIKEN Center for Advanced Intelligence Project (AIP), RIKEN, 103-0027 Tokyo, Japan. (e-mail: naoto.yokoya@riken.jp)}
\thanks{J. Chanussot is with the Univ. Grenoble Alpes, INRIA, CNRS, Grenoble INP, LJK, 38000 Grenoble, France, also with the Aerospace Information Research Institute, Chinese Academy of Sciences, 100094 Beijing, China. (e-mail: jocelyn@hi.is)}
\thanks{J. Xu is with the Remote Sensing Technology Institute (IMF), German Aerospace Center (DLR), 82234 Wessling, Germany. (e-mail: jian.xu@dlr.de)}
\thanks{X. Zhu is with the Remote Sensing Technology Institute (IMF), German Aerospace Center (DLR), 82234 Wessling, Germany, and Signal Processing in Earth Observation (SiPEO), Technical University of Munich (TUM), 80333 Munich, Germany. (e-mail: xiaoxiang.zhu@dlr.de)}
}

% The paper headers
\markboth{IEEE Transactions on Cybernetics,~Vol.~XX, No.~XX, ~XXXX,~2020}%
{Shell \MakeLowercase{\textit{et al.}}: Joint and Progressive Subspace Analysis (JPSA) with Spatial-Spectral Manifold Alignment for Semi-Supervised Hyperspectral Dimensionality Reduction}

\maketitle
\begin{abstract}
\textcolor{blue}{This is the pre-acceptance version, to read the final version please go to IEEE Transactions on Cybernetics on IEEE Xplore.} Conventional nonlinear subspace learning techniques (e.g., manifold learning) usually introduce some drawbacks in explainability (explicit mapping) and cost-effectiveness (linearization), generalization capability (out-of-sample), and representability (spatial-spectral discrimination). To overcome these shortcomings, a novel linearized subspace analysis technique with spatial-spectral manifold alignment is developed for a semi-supervised hyperspectral dimensionality reduction (HDR), called joint and progressive subspace analysis (JPSA). The JPSA learns a high-level, semantically meaningful, joint spatial-spectral feature representation from hyperspectral data by 1) jointly learning latent subspaces and a linear classifier to find an effective projection direction favorable for classification; 2) progressively searching several intermediate states of subspaces to approach an optimal mapping from the original space to a potential more discriminative subspace; 3) spatially and spectrally aligning manifold structure in each learned latent subspace in order to preserve the same or similar topological property between the compressed data and the original data. A simple but effective classifier, i.e., nearest neighbor (NN), is explored as a potential application for validating the algorithm performance of different HDR approaches. Extensive experiments are conducted to demonstrate the superiority and effectiveness of the proposed JPSA on two widely-used hyperspectral datasets: Indian Pines (92.98\%) and the University of Houston (86.09\%) in comparison with previous state-of-the-art HDR methods. The demo of this basic work (i.e., ECCV2018) is openly available at \url{https://github.com/danfenghong/ECCV2018_J-Play}.
\end{abstract}

\graphicspath{{figures/}}


\section{Introduction}
\IEEEPARstart{H}{yperspectral} (HS) data are often characterized by rich and diverse spectral information, which enables us to locate or identify targets more easily. However, their high dimensionality also raises some crucial issues that need to be carefully considered, including information redundancy, complex noise effects, need for large storage capacities and high performance computing, and the curse of dimensionality. A general way to address this problem is to compress the original data to a low-dimensional and highly-discriminative subspace with the preservation of the topological structure. In general, it is also referred to as dimensionality reduction (DR) or subspace learning (SL).

Over the past decade, SL techniques have been widely used in remote sensing data processing and analysis \cite{yuan2015hyperspectral,wang2016salient,lu2017hybrid,wang2018optimal,wu2019orsim,hong2020more,wang2018scene,wu2020fourier,gao2020spectral,hong2020invariant}, particularly hyperspectral dimensionality reduction (HDR) \cite{li2018discriminant}. Li \textit{et al.} \cite{li2012locality} carried out the HDR and classification by locally preserving neighborhood relations. In \cite{gao2017optimized}, spectral-spatial noise estimation can largely enhance the performance of dimensionality reduction, and the proposed method not only can extract high-quality features but also can well deal with nonlinear problems in hyperspectral image classification. Authors of \cite{huang2015dimensionality} introduced the sparseness property \cite{hong2018augmented} into the to-be-estimated subspace in order to better structure the low-dimensional embedding space. Rasti \textit{et al.} \cite{rasti2016hyperspectral} extracted the hyperspectral features in an unsupervised fashion using the orthogonal total Variation component analysis (OTVCA), yielding a smooth spatial-spectral HSI feature extraction. In \cite{Hong2017}, a spatial-spectral manifold (SSM) embedding was developed to compress the HS data into a more robust and discriminative subspace. Wang \textit{et al.} \cite{wang2018hierarchical} proposed to select representative features hierarchically by the means of random projection in an end-to-end neural network, which has shown the effectiveness in the large-scale data. Very recently, Huang \textit{et al.} \cite{huang2019dimensionality} followed the trail of drawbacks of spatial-spectral techniques, and fixed them by designing a new spatial-spectral combined distance to select spatial-spectral neighbors of each HS pixel more effectively. In the combined distance, the pixel-to-pixel distance measurement between two spectral signatures is converted to the weighted summation distance between spatially adjacent spaces of the two target pixels.

Despite the good performance of these methods in HDR, yet most of them only adhere to either the unsupervised or the supervised strategy, and fail to jointly consider the labeled and unlabeled information in the process of HDR. Some recent works for semi-supervised HDR have been proposed by the attempt to preserve the potentially global data structure that lies in the whole high-dimensional space. For example, Liao \textit{et al.} \cite{liao2013semisupervised} simultaneously exploited labeled and unlabeled data to extract the feature representation from the HSI in a semi-supervised fashion, called semi-supervised local discriminant analysis (SELD). Different from \cite{liao2013semisupervised} that utilizes the similarity measurement to construct the graph structure, in \cite{zhao2014general}, the performance of LDA is enhanced with the joint use of the labels and ``soft-labels'' predicted by label propagation, yielding a soft-label LDA (SLLDA) for semi-supervised HDR. A similar semi-supervised strategy was presented in \cite{wu2018semi} to reduce the spectral dimension of HSI by embedding pseudo-labels obtained using the pre-trained classifier into LFDA, called semi-supervised LFDA (SSLFDA). The use of ``soft-labels'' or ``pseudo-labels'' is effective for the process of low-dimensional embedding. Since more pixels considered can help us better capture the global manifold of the data, even though these soft or pseudo-labels could be noisy and inaccurate. It should be noted that these techniques are commonly applied as a disjunct feature learning step before classification, whose limitation mainly lies in a weak connection between features by SL and label space (see the top panel of Fig. 1). It is unknown which learned features can accurately improve the classification. In \cite{gao2020combining}, the features can adequately exploited by using the t-distributed stochastic neighbor embedding and a multi-scale scheme, and the proposed neural network shows outstanding and reliable performance in HS image classification.

\begin{figure}[!t]
\centering
        \includegraphics[width=0.5\textwidth]{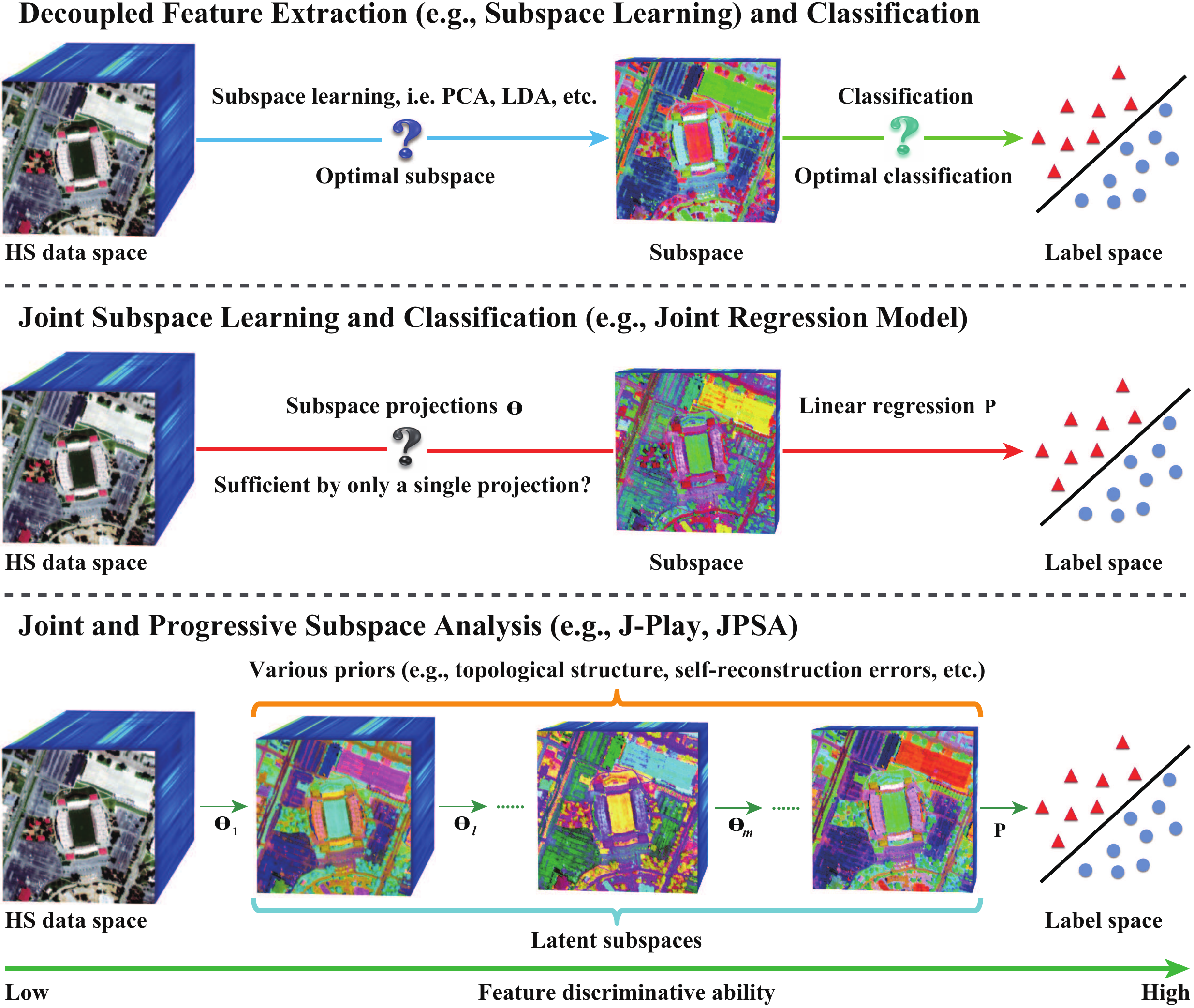}
\caption{The motivation interpolation from separately learning subspaces and training classifier \cite{Martnez2001}, to jointly learning subspaces and classifier \cite{ji2009linear}, to joint and progressive learning multi-coupled subspaces and classifier again \cite{hong2018joint}. The green bottom line from left to right indicates a gradual improvement in feature discriminative ability. Ideally, the features (subspaces) learned by our model are expected to have a higher discrimination ability, which benefits from the proposed joint and progressive learning strategy.}
\label{fig:Motivation}
\end{figure}

A feasible solution to this problem can be generalized into a joint learning framework \cite{ji2009linear} that simultaneously learns a linearized subspace projection and a classifier, as illustrated in the middle panel of Fig. 1. Inspired by it, a large amount of work has been proposed for various applications, such as cross-modality learning and retrieval \cite{hong2020learning}, and heterogeneous joint features learning \cite{hong2020x}. Although these works have tried to make a connection between the learned subspaces and label information using regression techniques (e.g., linear regression) to adaptively find a latent subspace in favor of classification, they fail to find an optimal subspace. It is that the representative ability only using a single linear projection remains limited for the complex transformation from the original data space to the potential optimal subspace. Similar to the joint learning model, deep neural networks (DNN) have attracted increasing attention due to its powerful ability in HS feature extraction. Chen \textit{et al.} \cite{chen2014deep} designed a stacked autoencoder (SAE) for feature extraction and classification of HSI. In \cite{kemker2017self}, the authors investigated the performance of self-taught feature learning (e.g., convolutional autoencoder (CAE)) by jointly considering the spatial-spectral information embedding with the application to HSI classification.

\subsection{Motivation and Objectives}
To sum up, these aforementioned methods can be approximately categorized into linear HDR and nonlinear HDR techniques. Consequently, the strengths and weaknesses of the two methods can be summarized as follows.

1) Theoretically, nonlinear HDR strategies, such as manifold learning \cite{hong2019learning} and DNN-based DR methods (e.g., SAE and CAE) \cite{rasti2020feature}, can over-fit the data perfectly, owing to their powerful model learning capability. However, this type of method is relatively sensitive to complex spectral variability inevitably caused by complex noise, atmospheric effects, and various physical and chemical factors in hyperspectral imaging. Because the spectral variability tends to be absorbed by the DNN-based methods \cite{hang2019cascaded}, the discriminative ability of dimension-reduced feature gets possibly hurt.

2) In turn, the linearized SL methods, such as principal component analysis (PCA) \cite{Wold1987}, linearized manifold learning (e.g., locality preserving projection (LPP) \cite{He2004}), local discriminant analysis (LDA) \cite{Martnez2001}, and local fisher discriminant analysis (LFDA) \cite{Sugiyama2007}) can well address the above drawbacks, yet they usually provide limited performance due to the defects of the model itself, that is, the single linearized model is lack of data representation ability.

The above trade-off motivates us to develop a multi-layered linearized SL technique for HDR with more discriminative and robust data representation and to preserve the structural consistency between the compressed data and the original data.

\begin{figure*}[!t]
\centering
        \includegraphics[width=0.95\textwidth]{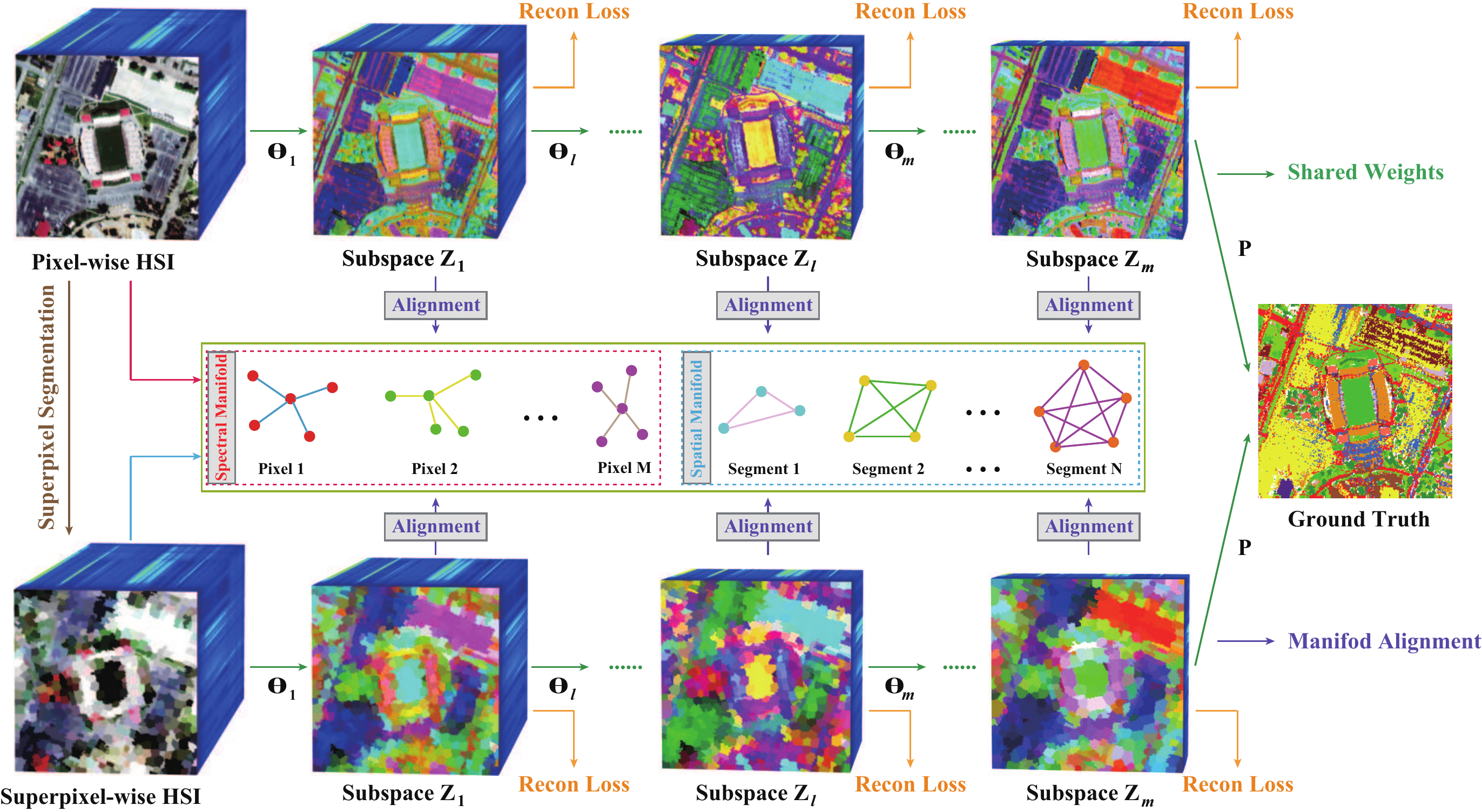}
\caption{The illustration of the proposed JPSA framework.}
\label{workflow}
\end{figure*}

\subsection{Method Overview and Contributions}
To effectively pursue high spectral discrimination and preservation of the spatial-spectral topological structure in compressing the HS data, we propose a novel \textbf{j}oint and \textbf{p}rogressive \textbf{s}ubspace \textbf{a}nalysis (JPSA) to linearly find an optimal subspace for the low-dimensional data representation, as shown in the bottom panel of Fig. \ref{fig:Motivation}. A promising idea of simultaneous SL and classification is used to form the basic skeleton of the proposed JPSA model. In the framework, we learn a series of subspaces instead of a single subspace, making the original data space being progressively converted to a potentially optimal subspace through multi-coupled intermediate transformations. To avoid trivial solutions, a self-reconstruction (SR) strategy in the form of regularization is applied in each latent subspace. Furthermore, we not only consider structure consistency (topology) between the compressed data and the original data in both spatial and spectral domains, but also align the two (spatial and spectral) manifolds in each latent subspace, yielding the SSM embedding in the process of HDR.

Beyond previous existing works, i.e., \cite{hong2018joint,hong2019cospace}, the main contributions of our work can be summarized as follow:
\begin{itemize}
\item We develop a novel semi-supervised HDR framework (JPSA) for better learning the spatial-spectral low-dimensional embedding by modeling relations between superpixels and pixels in a joint and progressive fashion.
\item With the SR term simultaneously performed on superpixels and pixels, the linearized JPSA shows its robustness and effectiveness in handling the spectral variability over many nonlinear HDR approaches, which will be well demonstrated in the following experiment section.
\item Spatial-spectral manifolds are preserved in each latent subspace and are further aligned for spatial-spectral structure consistency between the compressed data and the original data, where the manifold structure in spectral space is computed by Gaussian kernel function, and the spatial manifold structure is determined by superpixels, e.g., simple linear iterative clustering (SLIC)~\cite{achanta2012slic}.
\item To avoid falling into bad local optimums, a pre-training model, called \textbf{auto}-\textbf{r}econstructing \textbf{u}nsupervised \textbf{le}arning (AutoRULe), is proposed as an initialization of JPSA to jointly initialize the branches of pixels and superpixels.
\item An iterative optimization algorithm based on the alternating direction method of multipliers (ADMM) is designed to solve the newly-proposed model.
\end{itemize}

%The remainder of this paper is structured as follows. Section II details the joint and progressive subspace analysis (JPSA) and corresponding optimization algorithm. Section III reports the experimental results on two hyperspectral datasets and provides a qualitative and quantitative analysis. Finally, Section IV concludes with a summary.

\section{JPSA: Joint \& Progressive Subspace Analysis}

Fig. 2 illustrates the workflow of the proposed JPSA. Intuitively, the JPSA is a two-stream multi-layered regression model involving the two input sources: pixel-wise and superpixel-wise spectral signatures and the same output (ground truth). In the learning process of the two-stream model, the to-be-estimated parameters (projections) are shared with a spatial-spectral alignment constraint in each latent subspace. Moreover, each learned subspace is expected to be capable of projecting back to its former high-dimensional product, which is measured by the reconstruction loss.

\subsection{Review of Joint Regression}
Before introducing our JPSA, we first briefly introduce the basis of developing our method: a joint regression model \cite{ji2009linear}, in which SL and classification are simultaneously performed to reduce the gap between the estimated subspace and labels. This model has been proven to be effective in extracting the discriminative low-dimensional representation \cite{hong2019learnable}. Let $\mathbf{X}=\lbrack \mathbf{x}_{1},...,\mathbf{x}_{k},...,\mathbf{x}_{N}\rbrack \in\mathbb{R}^{d_{0}\times N}$ be a HS data matrix with $d_{0}$ bands by $N$ pixels, and $\mathbf{Y} \in\{0,1\}^{L \times N}$ be the one-hot encoded class matrix corresponding to labels, whose $k$th column is defined as $\mathbf{y}_{k}=\lbrack \mathbf{y}_{k1},...,\mathbf{y}_{kt},...,\mathbf{y}_{kL}\rbrack^{\T}\in\mathbb{R}^{L\times 1}$, we then have
\begin{equation}
\label{eq1}
\begin{aligned}
      \mathop{\min}_{\mathbf{P},\mathbf{\Theta}}\frac{1}{2}\norm{\mathbf{Y}-\mathbf{P}\mathbf{\Theta}\mathbf{X}}_{\F}^{2}+\frac{\alpha}{2}\norm{\mathbf{P}}_{\F}^{2} \; \; {\rm s.t.}\;\; \mathbf{\Theta}\mathbf{\Theta}^{\T} = \mathbf{I},
\end{aligned}
\end{equation}
where $\norm{\bullet}_{\F}$ represents a Frobenius norm; $\mathbf{P}\in\mathbb{R}^{L\times d_{m}}$ ($d_{m}$ denotes the dimension of the latent subspace) is regression matrix to explicitly bridge the learnt latent subspace and labels, and the projection $\mathbf{\Theta}\in\mathbb{R}^{d_{m}\times d_{0}}$  is usually called as intermediate transformation and the corresponding subspace $\mathbf{\Theta}\mathbf{X}$ is called as the latent subspace. It has been proven in \cite{Hu2016} that the feature is prone to be structurally learned and represented in such a latent subspace.

Further, by considering the graph structure measured by an adjacency matrix $\mathbf{W}\in\mathbb{R}^{N\times N}$ as a regularizor \cite{Yan2007_2}, the joint regression model in Eq. (\ref{eq1}) can be extended to the following improved version \cite{hong2019cospace}:
\begin{equation}
\label{eq2}
\begin{aligned}
\mathop{\min}_{\mathbf{P},\mathbf{\Theta}}\frac{1}{2}\norm{\mathbf{Y}-\mathbf{P}\mathbf{\Theta}\mathbf{X}}_{\F}^{2}&+\frac{\alpha}{2}\norm{\mathbf{P}}_{\F}^{2} +  \frac{\beta}{2}\tr(\mathbf{\Theta}\mathbf{X}\mathbf{L}\mathbf{X}^{\T}\mathbf{\Theta}^{\T})\\
&\mathrm{s.t.} \;\; \mathbf{\Theta}\mathbf{\Theta}^{\T}=\mathbf{I},
\end{aligned}
\end{equation}
where $\mathbf{D}_{ii}=\sum_{j}\mathbf{W}_{ij}$ is defined as a degree matrix and the Laplacian matrix $\mathbf{L}$ can be computed by $\mathbf{L}=\mathbf{D}-\mathbf{W}$ \cite{Chung1997}. The third term of Eq. (\ref{eq2}), i.e., graph regularization, can provide additional prior knowledge by modeling relations between samples, thereby improving the regression performance.

\subsection{Problem Formulation}
A single linear transformation is hardly capable of describing the complex mapping relationship between the data and labels well, particularly for HS data suffering from a variety of spectral variabilities. On the other hand, although the nonlinear techniques (e.g., manifold learning or DL) hold a powerful representation ability for the HS data, yet they are usually vulnerable to the attack of spectral variability, inevitably degrading the quality of dimension-reduced features. As a trade-off, we propose to progressively learn multi-coupled linear projections on the basis of the joint regression framework. Thus, the resulting JPSA with necessary priors can be formulated as the following constrained optimization problem:
\begin{equation}
\label{eq3}
\begin{aligned}
       \mathop{\min}_{\mathbf{P},\{\mathbf{\Theta}_{l}\}_{l=1}^{m}}&\frac{1}{2}\mathbf{\Upsilon}(\{\mathbf{\Theta}_{l}\}_{l=1}^{m})
       +\frac{\alpha}{2}\mathbf{E}(\mathbf{P},\{\mathbf{\Theta}_{l}\}_{l=1}^{m})+\frac{\beta}{2}\mathbf{\Phi}(\{\mathbf{\Theta}_{l}\}_{l=1}^{m})\\
       &+\frac{\gamma}{2}\mathbf{\Psi}(\mathbf{P})\\
       &\mathrm{s.t.} \;\; \mathbf{X}_{l}\succeq 0, \; \norm{\mathbf{x}_{lk}}_{2} \preceq 1, \; \mathbf{X}_{l}^{sp}\succeq 0, \; \norm{\mathbf{x}_{lk}^{sp}}_{2} \preceq 1,
\end{aligned}
\end{equation}
where $\{\mathbf{\Theta}_{l}\}_{l=1}^{m}\in\mathbb{R}^{d_{l}\times d_{l-1}}$ are defined as a set of intermediate transformations, $m$ is the number of subspace projections, and $\{d_{l}\}_{l=1}^{m}$ stand for as the dimensions of those latent subspaces. Moreover, $\mathbf{X}_{l}$ denotes the $l$-th layer subspace features, where $\mathbf{X}_{0}$ represents original data ($\mathbf{X}$), while $\mathbf{X}_{l}^{sp}$ denotes the superpixel representation of $\mathbf{X}_{l}$. To effectively solve the two-stream joint regression model in Eq. (\ref{eq3}), several key terms are featured in the following.

\emph{1) SR Loss Term $\mathbf{\Upsilon}(\{\mathbf{\Theta}_{l}\}_{l=1}^{m})$}: Without any constraints or prior, jointly estimating multiple successive variables in JPSA can hardly be implemented, especially when the number of estimated variables gradually increases. This can be well explained by gradient missing between the two neighboring variables estimated in the process of optimization. In other words, the variations between two neighboring variables approach to a tiny value or even zero. When the number of estimated projections accumulates to a certain extent, most of the valid values could only gather a few projections, making other projections being close to identity matrix and become meaningless. To address the issue mentioned above, a kind of autoencoder-like scheme is adopted to reduce the information loss in the process of propagation between two neighboring spaces. The benefits of the scheme are two-folds. On the one hand, this term can prevent over-fitting of the data to a great extent, especially avoiding all kinds of spectral variabilities from being considered, since we found that those variabilities are difficult to be linearly reconstructed. On the other hand, it can also establish an effective link between the original space and the subspace, enabling the learned subspace to project back to the former one as much as possible. Such a strategy can be formulated by simultaneously considering pixels and superpixels of HSI:
\begin{equation}
\label{eq4}
\begin{aligned}
      \mathbf{\Upsilon}(\{\mathbf{\Theta}_{l}\}_{l=1}^{m})=\sum_{l=1}^{m}\norm{[\mathbf{X}_{l-1}\;\;\mathbf{X}_{l-1}^{sp}]-\mathbf{\Theta}_{l}^{\T}\mathbf{\Theta}_{l}[\mathbf{X}_{l-1}\;\;\mathbf{X}_{l-1}^{sp}]}_{\F}^{2}.
\end{aligned}
\end{equation}
Please note that we propose to utilize Eq. (\ref{eq4}) in each latent subspace to maximize the advantages of this term.

\emph{2) Prediction Loss Term} $\mathbf{E}(\mathbf{P},\{\mathbf{\Theta}_{l}\}_{l=1}^{m})$: This term is to minimize the empirical risk between the original data and the label matrix through a set of subspace projections and a linear regression coefficient, which can be written as
\begin{equation}
\label{eq5}
\begin{aligned}
      \mathbf{E}(\mathbf{P},\{\mathbf{\Theta}_{l}\}_{l=1}^{m})=\norm{[\mathbf{Y}\;\;\mathbf{Y}]-\mathbf{P}\mathbf{\Theta}_{m}...\mathbf{\Theta}_{l}...\mathbf{\Theta}_{1}[\mathbf{X}\;\;\mathbf{X}^{sp}]}_{\F}^{2}.
\end{aligned}
\end{equation}
Theoretically, with the increase of the number of estimated subspaces, the variations between neighboring subspaces are gradually narrowed down to a very small range. In this case, such small variations can be approximately represented via a \emph{linear transformation}. This allows us to find a good solution in a simple way, especially for the non-convex model.

\emph{3) Alignment-based SSM Regularization} $\mathbf{\Phi}(\{\mathbf{\Theta}_{l}\}_{l=1}^{m})$: As introduced in \cite{hong2020graph}, manifold structure is an important prior for compressing high-dimensional data, which can effectively capture the intrinsic structure between samples. For this reason, we not only embed the locally spectral manifold structure computed between the pixels, but also model the non-local-like spatial manifolds constructed by superpixels. Therefore, the two graph structure can be formulated as
\begin{equation}
\label{eq6}
   \mathbf{W}_{i,j}^{p}=
    \begin{cases}
      \begin{aligned}
     {\rm exp}&{\frac{-||\mathbf{X}_{i}-\mathbf{ X}_{j}||_{2}^{2}}{2\sigma^{2}}}, \; \; \text{if \(\mathbf{X}_{j}\)\;\(\in\)\;\(\phi_{k}(\mathbf{X}_{i})\);}\\
      0, \;  \;& \text{otherwise,}
      \end{aligned}
    \end{cases}
\end{equation}
\begin{equation}
\label{eq7}
   \mathbf{W}_{i,j}^{sp}=
    \begin{cases}
      \begin{aligned}
     {\rm exp}&{\frac{-||\mathbf{X}_{i}^{sp}-\mathbf{ X}_{j}^{sp}||_{2}^{2}}{2\sigma^{2}}}, \; \; \text{if \(\mathbf{X}_{j}^{sp}\)\;\(\in\)\;\(\phi_{k}(\mathbf{X}_{i}^{sp})\);}\\
      0, \;  \;& \text{otherwise,}
      \end{aligned}
    \end{cases}
\end{equation}
where $\phi_{k}(\mathbf{X}_{i})$ and $\phi_{k}(\mathbf{X}_{i}^{sp})$ are the $k$ neighbors of the pixel $\mathbf{X}_{i}$ and the superpixel $\mathbf{X}_{i}^{sp}$, respectively.

Additionally, we also align the spatial-spectral manifolds in each learned subspace to enhance the model's ability to distinguish and generalize, further yielding the structure consistency of the two-stream joint regression model. The alignment operator can be expressed by the form of a graph:
\begin{equation}
\label{eq8}
   \mathbf{W}_{i,j}^{a}=
    \begin{cases}
      \begin{aligned}
      1, \; \;& \text{if \(\mathbf{X}_{i}\)\;\(\in\)\;\(\phi(\mathbf{X}_{j}^{sp})\);}\\
      0, \;  \;& \text{otherwise,}
      \end{aligned}
    \end{cases}
\end{equation}
where $\phi(\mathbf{X}_{j}^{sp})$ denotes the pixel set in the $j$-th superpixel.

\begin{figure}[!t]
	  \centering
		\subfigure{
			\includegraphics[width=0.4\textwidth]{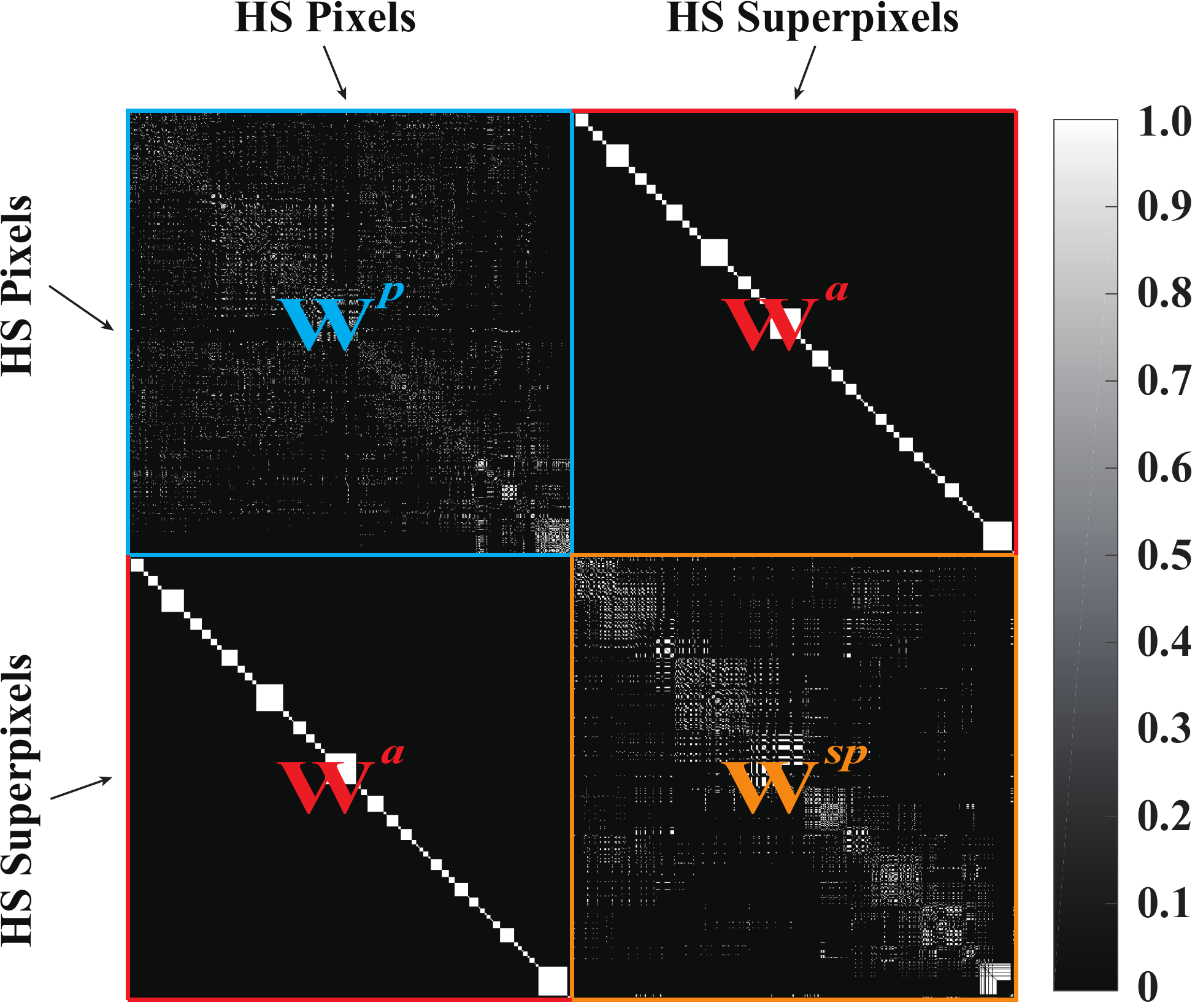}
		}
        \caption{A showcase to illustrate the graph structure used in the alignment-based SSM regularization term.}
\label{fig:graph}
\end{figure}

By collecting the above sub-graphs, we have the final graph structure ($\mathbf{W}^{f}$) by considering spatial and spectral neighbors of each pixel as well as their alignment information:
\begin{equation}
\label{eq9}
    \begin{aligned}
    	 \mathbf{W}^{f}=\left[
    	 \begin{matrix}
                \mathbf{W}^{p}&\mathbf{W}^{a}\\
    	        \mathbf{W}^{a}&\mathbf{W}^{sp}\\
         \end{matrix}
         \right].
    \end{aligned}
\end{equation}
Thus, the resulting manifold alignment-based spatial-spectral regularization can be written as
\begin{equation}
\label{eq10}
\begin{aligned}
      \mathbf{\Phi}(\{\mathbf{\Theta}_{l}\}_{l=1}^{m})=\sum_{l=1}^{m}\tr(\mathbf{\Theta}_{l}[\mathbf{X}_{l-1}\;\mathbf{X}_{l-1}^{sp}]\mathbf{L}^{f}[\mathbf{X}_{l-1}\;\mathbf{X}_{l-1}^{sp}]^{\T}\mathbf{\Theta}_{l}^{\T}),
\end{aligned}
\end{equation}
where $\mathbf{L}^{f}$ can be computed by $\mathbf{D}^{f}-\mathbf{W}^{f}$. In this study, each pixel's spatial neighbors are other pixels in the same segment obtained by SLIC, while its $k$ spectral neighbors are selected with Euclidean measurement on a kernel-induced space. Fig.~\ref{fig:graph} illustrates the spatial-spectral graph structure.

\emph{4) Regression Coefficient Regularization }$\mathbf{\Psi}(\mathbf{P})$: This regularization term ensures a reliable solution and improves the generalization ability of the model, which is
\begin{equation}
\label{eq11}
\begin{aligned}
      \mathbf{\Psi}(\mathbf{P})=\norm{\mathbf{P}}_{\F}^{2}.
\end{aligned}
\end{equation}

%HS data is acquired in a non-negative unit.
%This is due to the existing of physical meaning behind the non-negativity.
Hyperspectral data are non-negative either in radiance or reflectance. To inherit this physical nature, we expect to learn non-negative features with respect to each learned low-dimensional feature (e.g., $\{\mathbf{X}_{l}\}_{l=1}^{m} \succeq 0$). The hard orthogonal constraint with respect to the variable $\mathbf{\Theta}$ could lead to non-convergence of the model or reach a bad solution. To provide a proper search space of the solution, we, therefore, relax the constraint by imposing a sample-based norm constraint \cite{Lee2007} on each latent subspace as $\norm{\mathbf{x}_{lk}}_{2} \preceq 1, \forall k=1,...,N$ and $l=1,...,m$. Note that these constraints are similarly applicable to the superpixel-guided optimization problem.

\begin{figure*}[!t]
	  \centering
		\subfigure[Indian Pines Dataset]{
			\includegraphics[width=0.95\textwidth]{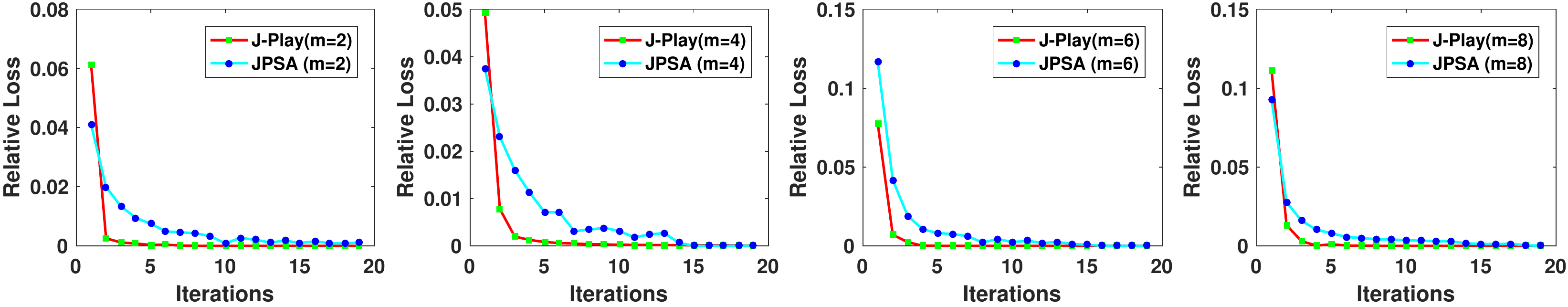}
            \label{fig:Convergence_Indian}
		}
		\subfigure[University of Houston Dataset]{
			\includegraphics[width=0.95\textwidth]{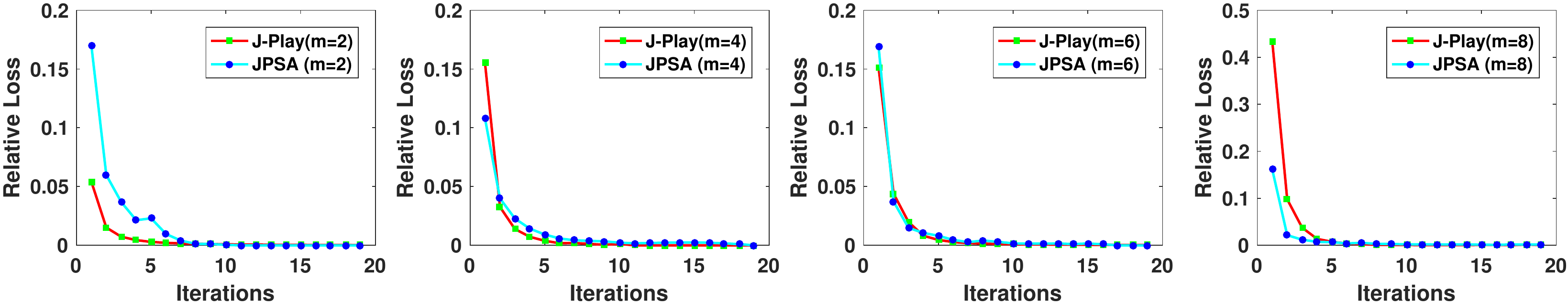}
            \label{fig:Convergence_Houston}
		}
         \caption{Convergence analysis of J-Play and JPSA with different $m$ values of $2,4,6,8$ (left to right) was experimentally performed on the two HS datasets. (a): Indian Pines Dataset. (b): University of Houston Dataset.}
\label{fig:Convergence}
\end{figure*}

\begin{algorithm}[!t]
\small
\caption{JPSA: global algorithm}
\KwIn{$\mathbf{Y}, \mathbf{\tilde{X}}, \mathbf{L}^{f},$ and parameters $\alpha, \beta, \gamma$ and $maxIter.$}
\KwOut{$\{\mathbf{\Theta}_{l}\}_{l=1}^{m}.$}
{$\textbf{Initialization Step: }$}\\
{Greedily initialize $\mathbf{\Theta}_{l}$ corresponding to each latent subspace}:\\
\For{$l=1:m$}
{
  $\mathbf{\Theta}^{0}_{l} \leftarrow LPP(\mathbf{\tilde{X}}_{l-1})$\\
  $\mathbf{\Theta}_{l} \leftarrow AutoRULe(\mathbf{\tilde{X}}_{l-1},\mathbf{\Theta}^{0}_{l},\mathbf{L}^{f})$\\
  $\mathbf{\tilde{X}}_{l} \leftarrow \mathbf{\Theta}_{l}\mathbf{\tilde{X}}_{l-1}$\\
}
{$\textbf{Fine-tuning Step: }$}\\
  $t=0,\zeta=1e-4;$\\
  \While{$t>maxIter$}
 {
  % Fix other variables to update $\mathbf{P}$ by solving a subproblem $\mathbf{P}$;\\
  Update $\mathbf{P}$ by solving a subproblem in Eq. (\ref{eq16}).\\
   \For{$i=1:m$}
   {
      %Fix other variables to update $\mathbf{\Theta}_{l}^{t+1}$ by solving a subproblem of $\mathbf{\Theta}_{l};$
      Update $\mathbf{\Theta}_{l}^{t+1}$ by solving a subproblem in Eq. (\ref{eq18}).
   }
  Compute the objective function value $Obj^{t+1}$ and check the convergence condition:\\
   \eIf{$|\frac{Obj^{t+1}-Obj^{t}}{Obj^{t}}|<\zeta$}
   {
     Stop iteration;
   }
   {
     $t\leftarrow t+1$;
   }
 }
\end{algorithm}

\subsection{Model Learning}
Considering the fact that we need to successively estimate multi-coupled variables in JPSA, which obviously results in the increasing complexity and the non-convexity of our model, a group of good initial approximations of subspace projections $\{\mathbf{\Theta}_{l}\}_{l=1}^{m}$ would greatly reduce training time and help finding a better local optimal solution. This is a common tactic that has been widely used to address this issue \cite{Hinton2006}. Inspired by this trick, we pre-train our model by simplifying Eq.(\ref{eq3}) as
\begin{equation}
\label{eq12}
\begin{aligned}
      \mathop{\min}_{\mathbf{\Theta}_{l}}\frac{1}{2}\mathbf{\Upsilon}(\mathbf{\Theta}_{l})+\frac{\eta}{2}\mathbf{\Phi}(\mathbf{\Theta}_{l}) \;\; \mathrm{s.t.} \;\; \mathbf{\tilde{X}}_{l}\succeq 0, \; \norm{\mathbf{\tilde{x}}_{lk}}_{2} \preceq 1,
\end{aligned}
\end{equation}
where $[\mathbf{X}_{l}\;\;\mathbf{X}_{l}^{sp}]$ is collectively rewritten as $\mathbf{\tilde{X}}_{l}$ for convenience of writing and model optimization.

We call the Eq. (\ref{eq12}) as \textbf{auto}-\textbf{r}econstructing \textbf{u}nsupervised \textbf{le}arning (AutoRULe). Given the outputs of AutoRULe to the problem of Eq. (\ref{eq3}) as the initialization, $\{\mathbf{\Theta}_{l}\}_{l=1}^{m}$ and $\mathbf{P}$ tend to obtain the better estimations. In details, \textbf{Algorithm 1} summarizes the global algorithm of JPSA, where AutoRULe is initialized by LPP.

We propose to use the ADMM-based optimization method to solve the pre-training method (AutoRULe), hence an equivalent form of Eq. (\ref{eq12}) is considered by introducing multiple auxiliary variables $\mathbf{H}$, $\mathbf{G}$, $\mathbf{Q}$ and $\mathbf{S}$ to replace $\mathbf{\tilde{X}}_{l}$, $\mathbf{\Theta}_{l}$, $\mathbf{\tilde{X}}_{l}^{+}$ and $\mathbf{\tilde{X}}_{l}^{\sim}$, respectively, where $()^{+}$ denotes an operator for converting each component of the matrix to its absolute value and $()^{\sim}$ is a proximal operator that solves the constraint of $\norm{\mathbf{\tilde{x}}_{lk}}_{2} \preceq 1$ \cite{Heide2015}. Therefore, the resulting expression is
\begin{equation}
\label{eq13}
\begin{aligned}
       \mathop{\min}_{\mathbf{\Theta}_{l},\mathbf{H},\mathbf{G},\mathbf{Q},\mathbf{S}}&\frac{1}{2}\norm{\mathbf{\tilde{X}}_{l-1}-\mathbf{G}^{\T}\mathbf{H}}_{\F}^{2}+\frac{\eta}{2}\tr(\mathbf{\Theta}_{l}\mathbf{\tilde{X}}_{l-1}\mathbf{L}^{f}\mathbf{\tilde{X}}_{l-1}^{\T}\mathbf{\Theta}_{l}^{\T})\\
       &\mathrm{s.t.} \;\; \mathbf{\tilde{X}}_{l}=\mathbf{\Theta}_{l}\mathbf{\tilde{X}}_{l-1}, \;\mathbf{Q}\succeq 0, \;\norm{\mathbf{s}_{k}}_{2} \preceq 1.\\
        &\qquad \mathbf{\tilde{X}}_{l}=\mathbf{H}=\mathbf{Q}=\mathbf{S},\; \mathbf{\Theta}_{l}=\mathbf{G}.
\end{aligned}
\end{equation}
The constrained optimization problem in Eq. (\ref{eq13}) can be converted to its augmented Lagrangian version by introducing the Lagrange multipliers $\{\mathbf{\Lambda}_{n}\}_{n=1}^{4}$ and the penalty parameter $\mu$, where the non-negativity and norm constraint can be relaxed by defining two kinds of proximal projection operators $l_{R}^{+}(\bullet)$ and $l_{R}^{\sim}(\bullet)$. More specifically, $l_{R}^{+}(\bullet)$ can be expressed as
\begin{equation}
\label{eq14}
\begin{aligned}
      max(\bullet)=
       \begin{cases}
       \hspace{0.15cm} \bullet \hspace{0.12cm}, \hspace{0.15cm} \bullet \succ 0\\
       \hspace{0.15cm} 0 \hspace{0.12cm}, \hspace{0.15cm}  \bullet \preceq 0,
       \end{cases}
\end{aligned}
\end{equation}
while $l_{R}^{\sim}(\bullet_{k})$ is a sample-based normalization operator:
\begin{equation}
\label{eq15}
\begin{aligned}
      prox_{f}(\bullet_{k})=
       \begin{cases}
       \hspace{0.15cm} \frac{\bullet_{k}}{\norm{\bullet_{k}}_{2}} \hspace{0.12cm}, \hspace{0.15cm} \norm{\bullet_{k}}_{2} \succ 1\\
       \hspace{0.45cm} \bullet_{k} \hspace{0.3cm}, \hspace{0.15cm}  \norm{\bullet_{k}}_{2} \preceq 1,
       \end{cases}
\end{aligned}
\end{equation}
where $\bullet_{k}$ is the $k$-th column of matrix $\bullet$ in our case.

\begin{algorithm}[!t]
\small
\caption{AutoRULe: initialization step for JPSA}
\KwIn{$\mathbf{\tilde{X}}_{l-1},\mathbf{\Theta}^{0}_{l},\mathbf{L}^{f},$ and parameters $\eta$ and $maxIter$.}
\KwOut{$\mathbf{\Theta}_{l}.$}
\textbf{Initialization}: $\mathbf{H}^{0}=\mathbf{\Theta}^{0}_{l}\mathbf{\tilde{X}}_{l-1}, \mathbf{G}^{0}=\mathbf{0}, \mathbf{Q}^{0}=\mathbf{P}^{0}=\mathbf{0}, \mathbf{\Lambda}_{2}^{0}=\mathbf{0}, \mathbf{\Lambda}_{1}^{0}=\mathbf{\Lambda}_{3}^{0}=\mathbf{\Lambda}_{4}^{0}=\mathbf{0}, \mu^{0}=1e-3, \mu_{max}=1e6, \rho=2, \varepsilon=1e-6, t=0.$\\
  \While{$t>maxIter$}
 {
         Fix $\mathbf{H}^{t}, \mathbf{G}^{t}, \mathbf{Q}^{t}, \mathbf{P}^{t}$ to update $\mathbf{\Theta}^{t+1}_{l}$ by Eq. (\ref{eq22}).\\
         Fix $\mathbf{\Theta}^{t+1}_{l}, \mathbf{G}^{t}, \mathbf{Q}^{t}, \mathbf{P}^{t}$ to update $\mathbf{H}^{t+1}$ by Eq. (\ref{eq24}).\\
         Fix $\mathbf{H}^{t+1}, \mathbf{\Theta}^{t+1}_{l}, \mathbf{Q}^{t}, \mathbf{P}^{t}$ to update $\mathbf{G}^{t+1}$ by Eq. (\ref{eq26}).\\
         Fix $\mathbf{H}^{t\!+\!1}, \mathbf{G}^{t\!+\!1}, \mathbf{\Theta}^{t\!+\!1}_{l}, \mathbf{P}^{t}$ to update $\mathbf{Q}^{t+1}$ by Eq. (\ref{eq28}).\\
         Fix $\mathbf{H}^{t\!+\!1}\!,\! \mathbf{G}^{t\!+\!1}, \mathbf{\Theta}^{t\!+\!1}_{l}, \mathbf{Q}^{t\!+\!1}$ to update $\mathbf{P}^{t\!+\!1}$ by Eq. (\ref{eq30}).\\
         Update Lagrange multipliers using Eq. (\ref{eq31}).\\
         Update penalty parameter using  $\mu^{t+1}=min(\rho\mu^{t},\mu_{max})$.\\
         Check the convergence conditions:\\
         \eIf{$\norm {\mathbf{H}^{t+1}-\mathbf{\Theta}_{l}^{t+1}\mathbf{\tilde{X}}_{l-1}}_{\F}<\varepsilon$ and $\norm {\mathbf{G}^{t+1}-\mathbf{\Theta}_{l}^{t+1}}_{\F}<\varepsilon$ and $\norm {\mathbf{Q}^{t+1}-\mathbf{\Theta}_{l}^{t+1}\mathbf{\tilde{X}}_{l-1}}_{\F}<\varepsilon$ and $\norm {\mathbf{P}^{t+1}-\mathbf{\Theta}_{l}^{t+1}\mathbf{\tilde{X}}_{l-1}}_{\F}<\varepsilon$}
         {
           Stop iteration;
         }
         {
         $t\leftarrow t+1$;
         }
 }
\end{algorithm}

\textbf{Algorithm 2} lists the optimization procedures of AutoRULe, and the solution to each subproblem is detailed in \emph{Appendix A}.

After running the AutoRULe, its outputs can be fed into JPSA for the model initialization, and then the two subproblems (solve $\mathbf{P}$ and $\{\mathbf{\Theta}_{l}\}_{l=1}^{m}$) in Eq. (\ref{eq3}) can be optimized alternatively as follows:

\emph{Optimization with respect to $\mathbf{P}$ subproblem}: Typically, this is a Tikhonov-regularized least square regression problem, which can be formulated as
\begin{equation}
\label{eq16}
\begin{aligned}
       \mathop{\min}_{\mathbf{P}}\frac{\alpha}{2}\norm{\mathbf{\tilde{Y}}-\mathbf{P}\mathbf{\Theta}_{m}...\mathbf{\Theta}_{l}...\mathbf{\Theta}_{1}\mathbf{\tilde{X}}}_{\F}^{2}
       +\frac{\gamma}{2}\norm{\mathbf{P}}_{\F}^{2},
\end{aligned}
\end{equation}
where the variable $\mathbf{\tilde{Y}}$ is a collection of $[\mathbf{Y}\;\;\mathbf{Y}]$ similar to the variable $\mathbf{\tilde{X}}$. Intuitively, the analytical solution of Eq. (\ref{eq16}) can be directly derived as
\begin{equation}
\label{eq17}
\begin{aligned}
       \mathbf{P} \leftarrow (\alpha\mathbf{\tilde{Y}}\mathbf{V}^{\T})(\alpha\mathbf{V}\mathbf{V}^{\T}+\gamma\mathbf{I})^{-1},
\end{aligned}
\end{equation}
where $\mathbf{V}$ is assigned to $\mathbf{\Theta}_{m}...\mathbf{\Theta}_{l}...\mathbf{\Theta}_{1}\mathbf{\tilde{X}}, \forall l=1,...,m$.

\emph{Optimization with respect to $\{\mathbf{\Theta}_{l}\}_{l=1}^{m}$}: When other variables are fixed, the variable $\mathbf{\Theta}_{l}$ can be individually solved, hence the optimization problem for any $\mathbf{\Theta}_{l}$ can be written in the following general form:
\begin{equation}
\label{eq18}
\begin{aligned}
       \mathop{\min}_{\mathbf{\Theta}_{l}}&\frac{1}{2}\norm{\mathbf{\tilde{X}}_{l-1}-\mathbf{\Theta}_{l}^{\T}\mathbf{\Theta}_{l}\mathbf{\tilde{X}}_{l-1}}_{\F}^{2}+\frac{\alpha}{2}\norm{\mathbf{\tilde{Y}}-\mathbf{P}\mathbf{\Theta}_{m}\dots\mathbf{\Theta}_{1}\mathbf{\tilde{X}}}_{\F}^{2}\\
       &+\frac{\beta}{2}\tr(\mathbf{\Theta}_{l}\mathbf{\tilde{X}}_{l-1}\mathbf{L}^{f}\mathbf{\tilde{X}}_{l-1}^{\T}\mathbf{\Theta}_{l}^{\T})\\
       &\mathrm{s.t.} \;\; \mathbf{\tilde{X}}_{l}=\mathbf{\Theta}_{l}\mathbf{\tilde{X}}_{l-1}, \; \mathbf{\tilde{X}}_{l}\succeq 0, \; \norm{\mathbf{\tilde{x}}_{lk}}_{2} \preceq 1.
\end{aligned}
\end{equation}
Likewise, the problem of  Eq. (\ref{eq18}) can basically be solved by following the framework of \textbf{Algorithm 2} (More details regarding the variable optimization can be found in \textit{Appendix A}.). The only difference lies in the optimization of subproblem with respect to $\mathbf{H}$. Herein, we supplement the optimization problem of the variable $\mathbf{H}$ as follows
\begin{equation}
\label{eq19}
\begin{aligned}
      \mathop{\min}_{\mathbf{H}}&\frac{1}{2}\norm{\mathbf{\tilde{X}}_{l-1}-\mathbf{G}^{\T}\mathbf{H}}_{\F}^{2}+\frac{\alpha}{2}\norm{\mathbf{\tilde{Y}}-\mathbf{P}_{l}\mathbf{H}}_{\F}^{2}\\
      &+\mathbf{\Lambda}_{1}^{\T}(\mathbf{H}-\mathbf{\Theta}_{l}\mathbf{\tilde{X}}_{l-1})+\frac{\mu}{2}\norm{\mathbf{H}-\mathbf{\Theta}_{l}\mathbf{\tilde{X}}_{l-1}}_{\F}^{2}\\
      &\mathrm{s.t.} \;\; \mathbf{P}_{l}=\mathbf{P}_{l-1}\mathbf{\Theta}_{l+1}, \; \mathbf{P}_{0}=\mathbf{P},
\end{aligned}
\end{equation}
whose analytical solution is given by
\begin{equation}
\label{eq20}
\begin{aligned}
       \mathbf{H} \leftarrow &(\alpha\mathbf{P}_{l}^{\T}\mathbf{P}_{l}+\mathbf{G}\mathbf{G}^{\T}+\mu\mathbf{I})^{-1}\\
       &\times(\alpha\mathbf{P}_{l}^{\T}\mathbf{\tilde{Y}}+\mathbf{G}\mathbf{\tilde{X}}_{l-1}+\mu\mathbf{\Theta}_{l}\mathbf{\tilde{X}}_{l-1}-\mathbf{\Lambda}_{1}).
\end{aligned}
\end{equation}
Finally, the aforementioned optimization procedures are repeated until a stopping criterion is satisfied. % Please find more derivation for each subproblem of JPSA in \textit{Appendix B}.
\begin{table}[!t]
\centering
\caption{Scene categories, the number of training (TR) and test (TE) samples for each class on the two datasets: Indian Pines and University of Houston.}
\resizebox{0.49\textwidth}{!}{
\begin{tabular}{c||c|c|c||c|c|c}
\toprule[1.5pt]
\multirow{2}{*}{No.}&\multicolumn{3}{c||}{Indian Pines Dataset}&\multicolumn{3}{c}{University of Houston Dataset}\\
\cline{2-7} & Class Name & TR & TE & Class Name & TR & TE \\
\hline \hline 1&CornNotill&50&1384&HealthyGrass&198&1053\\
 2&CornMintill&50&784&StressedGrass&190&1064\\
 3&Corn&50&184&Synthetic Grass&192&505\\
 4&GrassPasture&50&447&Tree&188&1056\\
 5&GrassTrees&50&697&Soil&186&1056\\
 6&HayWindrowed&50&439&Water&182&143\\
 7&SoybeanNotill&50&918&Residential&196&1072\\
 8&SoybeanMintill&50&2418&Commercial&191&1053\\
 9&SoybeanClean&50&564&Road&193&1059\\
 10&Wheat&50&162&Highway&191&1036\\
 11&Woods&50&1244&Railway&181&1054\\
 12&BuildingsGrassTrees&50&330&Parking Lot1&192&1041\\
 13&StoneSteelTowers&50&45&Parking Lot2&184&285\\
 14&Alfalfa&15&39&Tennis Court&181&247\\
 15&GrassPastureMowed&15&11&Running Track&187&473\\
 16&Oats&15&5&--&--&--\\
\hline \hline &Total&695&9671&Total&2832&12197\\
\bottomrule[1.5pt]
\end{tabular}
}
\label{Table:TRTE}
\end{table}
\subsection{Convergence Analysis}
The iterative alternating strategy used in \textbf{Algorithm~1} is nothing but a block coordinate descent, whose convergence is theoretically guaranteed as long as each subproblem of Eq. (\ref{eq12}) is exactly minimized~\cite{bertsekas1999nonlinear}. Each subproblem optimized in \textbf{Algorithm~2} is strongly convex, and thus the ADMM-based optimization strategy can converge to a unique minimum when the parameters are updated in finite steps~\cite{boyd2011distributed,kang2020learning}. Moreover, we experimentally illustrate to clarify the convergences of J-Play and the proposed JPSA on the two HS datasets, where the relative errors of objective function value are recorded in each iteration (see Fig. \ref{fig:Convergence}).

\section{Experiments}
\subsection{Description of the Data}
The experiments are performed on two different standard HS datasets, corresponding to different contexts, different sensors, and different resolutions.

\begin{figure*}[!t]
	  \centering
		{
	     \includegraphics[width=0.88\textwidth]{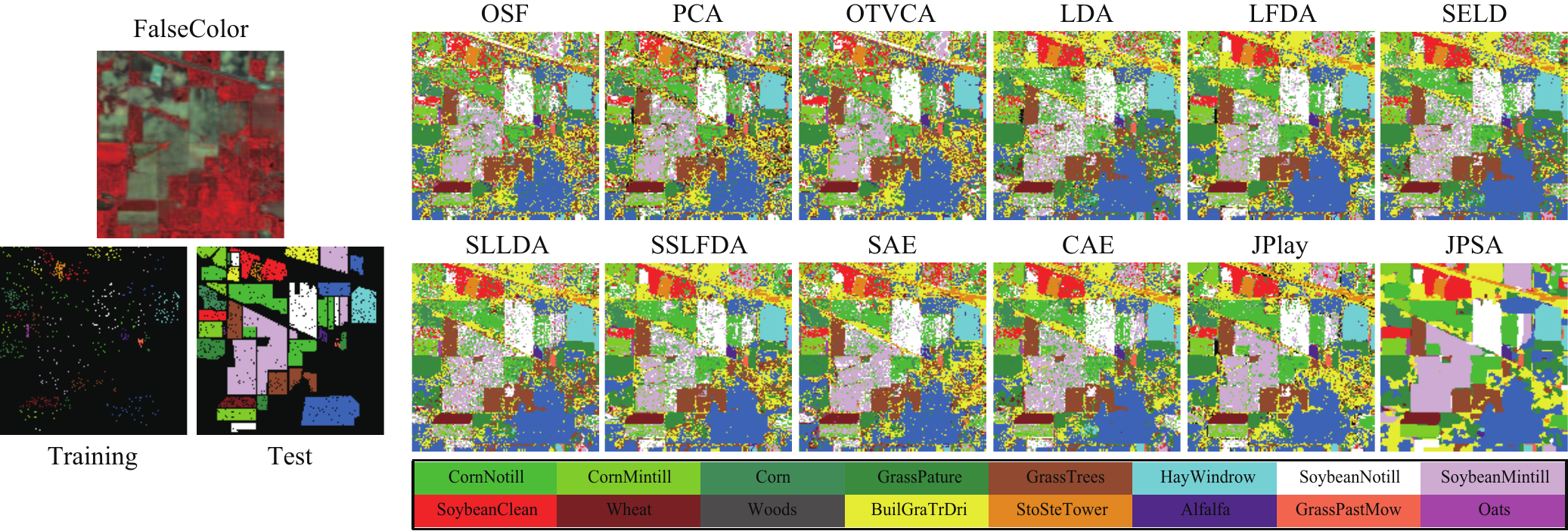}
		}	
         \caption{A false color image, the distribution of training and test sets with category names, and classification maps of the different algorithms obtained using the NN classifier on the Indian Pines dataset.}
\label{fig:CM_Indian}
\end{figure*}

1) \emph{Indian Pines AVIRIS Image:} The first HS cube was acquired by the AVIRIS sensor with $16$ classes of vegetation. It consists of $145\times145$ with the spectral 220 bands covering the wavelength range from 400nm to 2500nm in a 10nm spectral resolution. A set of widely-used training and test sets \cite{hong2018joint} with the specific categories is listed in Table \ref{Table:TRTE}. A false-color image of the data is given in Fig. \ref{fig:CM_Indian}.

\begin{figure*}[!t]
	  \centering
		{
	     \includegraphics[width=0.9\textwidth]{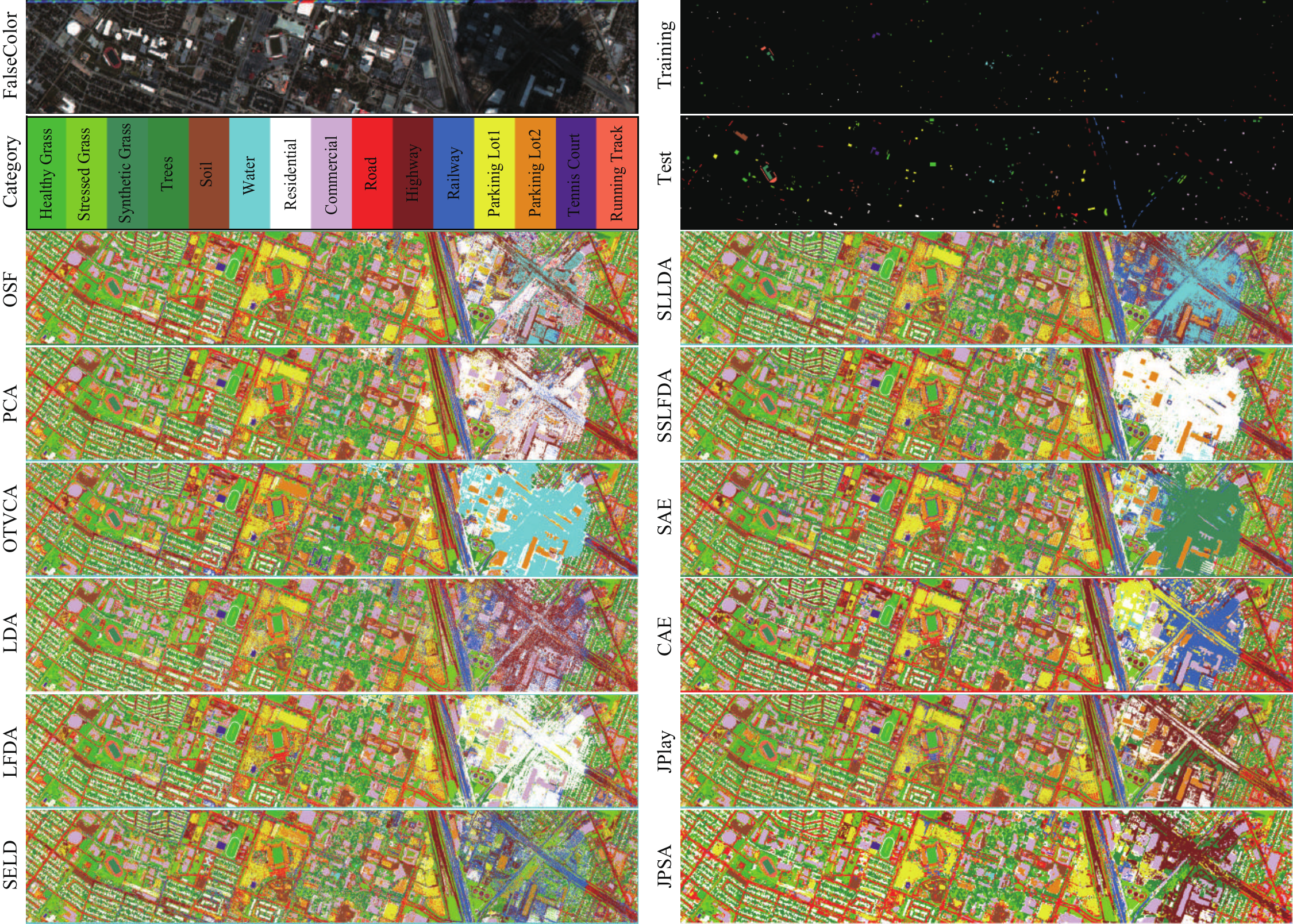}
		}	
         \caption{A false color image, the distribution of training and test sets with category names, and classification maps of the different algorithms obtained using the NN classifier on the University of Houston dataset.}
\label{fig:CM_Houston}
\end{figure*}

2) \emph{University of Houston Image:} The second HSI was provided for the $2013$ IEEE GRSS data fusion contest. It was acquired by an ITRES-CASI-1500 sensor over the campus of the University of Houston, Houston, USA, with a size of $349\times1905\times144$ in the wavelength from 364nm to 1046nm. The information regarding classes as well as training and test samples can be also found in Table \ref{Table:TRTE}. The first image of Fig. \ref{fig:CM_Houston} shows a false color image of the study scene.

\subsection{Experimental Setup and Preparation}
We learn the subspaces for the different methods on the training set and then the test set can be simply projected to the subspace where training and test samples can be further classified by the nearest neighbor (NN). The reason for selecting the simple but effective classifier in our case is that the NN classifier tends to avoid the confusing evaluation, as it is unknown whether the performance improvement originates from either the classifiers or the features themselves if more advanced classifiers are used.

Moreover, the original spectral features (OSF) without dimensionality reduction and ten popular and advanced methods are compared with our JPSA, including
\begin{itemize}
    \item Unsupervised HDR: PCA\cite{Wold1987}, OTVCA \cite{rasti2016hyperspectral};
    \item Supervised HDR: LDA \cite{Martnez2001}, LFDA \cite{sugiyama2007dimensionality}, and J-Play \cite{hong2018joint};
    \item Semi-supervised HDR: SELD \cite{liao2013semisupervised}, SLLDA \cite{zhao2014general}, and SSLFDA \cite{wu2018semi};
    \item DNN-based HDR: SAE \cite{chen2014deep}, CAE \cite{kemker2017self}.
\end{itemize}

Furthermore, we maximize the performances of the different algorithms by tuning their parameters, such as dimension ($d$), regularization parameters ($\alpha, \beta, \gamma$), etc., using 10-fold cross-validation on training data. Regarding the dimensions ($\{d_{l}\}$) which are common parameters for all algorithms, they can be selected ranging from $10$ to $50$ at an interval of $10$. For the number of nearest neighbors ($k$) and the standard deviation of Gaussian kernel function ($\sigma$) in those algorithms that need to construct the graph structure (e.g., LFDA, SELD, SSLFDA, J-Play, and JPSA), we select them in the range of $\{10, 20,...,50\}$ and $\{10^{-2}, 10^{-1}, \allowbreak 10^{0}, 10^{1}, 10^{2}\}$, respectively, and three regularization parameters ($\alpha, \beta, \gamma$) in J-Play or JPSA are all chosen from $\{10^{-2}, 10^{-1}, \allowbreak 10^{0}, 10^{1}, 10^{2}\}$. For the OTVCA algorithm, we directly used the parameter setting suggested in \cite{rasti2016hyperspectral}: that is, $d$ is equal to the number of classes, and $\lambda$ can be automatically determined by 1\% of the maximum intensity range of the datasets.

We adopt three criteria to quantitatively assess the algorithm performance, including \emph{Overall Accuracy} (OA), \emph{Average Accuracy} (AA), and  \emph{Kappa Coefficient} ($\kappa$). They can be formulated by using the following equations.
\begin{equation}
\begin{aligned}
      OA = \frac{N_{c}}{N_{a}},
\end{aligned}
\end{equation}
\begin{equation}
\begin{aligned}
      AA = \frac{1}{C}\sum_{i=1}^{C}\frac{N_{c}^{i}}{N_{a}^{i}},
\end{aligned}
\end{equation}
and
\begin{equation}
\begin{aligned}
      \kappa = \frac{OA-P_{e}}{1-P_{e}},
\end{aligned}
\end{equation}
where $N_{c}$ and $N_{a}$ denote the number of samples classified correctly and the number of total samples, respectively, while $N_{c}^{i}$ and $N_{a}^{i}$ correspond to the $N_{c}$ and $N_{a}$ of each class, respectively. $P_{e}$ in $\kappa$ is defined as the hypothetical probability of chance agreement \cite{cohen1960coefficient}, which can be computed by
\begin{equation}
\begin{aligned}
      P_{e} = \frac{N_{r}^{1}\times N_{p}^{1} + \dots N_{r}^{i}\times N_{p}^{i} + \dots + N_{r}^{C}\times N_{p}^{C}}{N_{a}\times N_{a}},
\end{aligned}
\end{equation}
where $N_{r}^{i}$ and $N_{p}^{i}$ denote the number of real samples for each class and the number of predicted samples for each class, respectively.

\begin{table*}[!t]
\centering
\caption{Quantitative performance comparisons of different algorithms on the Indian Pines dataset with the optimal parameter combination in terms of OA, AA, and $\kappa$, as well as the accuracy for each class. The best one is shown in bold. JPlay$_{4}$ means a four-layered J-Play model ($m=4$), while JPSA$_{4}$ denotes a four-layered JPSA model ($m=4$).}
\resizebox{0.85\textwidth}{!}{
\begin{tabular}{p{30pt}<{\centering}||p{30pt}<{\centering}|p{30pt}<{\centering}|p{32pt}<{\centering}||p{30pt}<{\centering}|p{30pt}<{\centering}||p{30pt}<{\centering}|p{30pt}<{\centering}|p{35pt}<{\centering}||p{30pt}<{\centering}|p{30pt}<{\centering}||p{30pt}<{\centering}|p{30pt}<{\centering}}
%\begin{tabular}{c||ccc|cc|ccc|cc|cc}
\toprule[1.5pt]
Method & OSF & PCA & OTVCA & LDA & LFDA & SELD & SLLDA & SSLFDA & SAE & CAE & JPlay$_{4}$ & JPSA$_{4}$\\
\hline \hline
$d$ & $220$ & $20$ & $16$ & $15$ & $15$ & $15$ & $15$ & $15$ & $20$ & $20$ & $20$ & $20$\\
$k$ & $-$ & $-$ & $-$ & $-$ & $10$ & $10$ & $-$ & $5$ & $-$ & $-$ & $10$ & $10$ \\
$\sigma$ & $-$ & $-$ & $-$ & $-$ & $0.1$ & $0.01$ & $-$ & $0.1$ & $-$ & $-$ & $0.1$ & $0.1$ \\
$\alpha$ & $-$ & $-$ & $-$ & $-$ & $-$ & $-$ & $-$ & $-$ & $-$ & $-$ & $1$ & $1$\\
$\beta$ & $-$ & $-$ & $-$ & $-$ & $-$ & $-$ & $-$ & $-$ & $-$ & $-$ & $0.1$ & $0.1$\\
$\gamma$ & $-$ & $-$ & $-$ & $-$ & $-$ & $-$ & $-$ & $-$ & $-$ & $-$ & $0.1$ & $0.1$\\
\hline\hline
OA&65.89&65.40&68.87&64.14&73.86&75.81&70.93&75.26&71.39&76.89&83.92&\bf92.98\\
AA&75.71&75.43&79.04&74.52&85.59&83.37&82.20&85.91&78.88&84.94&89.35&\bf95.40\\
$\kappa$&0.6148&0.6097&0.6490&0.5964&0.7042&0.7265&0.6713&0.7200&0.6765&0.7379&0.8169&\bf0.9197\\
\hline \hline
Class1&51.66&50.79&54.55&51.45&67.77&72.40&57.73&70.23&60.62&66.47&79.05&\bf91.04\\
Class2&57.40&57.14&59.69&48.47&65.05&65.69&59.69&67.35&56.51&72.19&80.74&\bf90.18\\
Class3&70.65&69.02&69.57&69.57&83.15&83.15&71.74&87.50&82.07&86.96&85.87&\bf99.46\\
Class4&88.14&87.92&90.60&90.60&\bf95.30&\bf95.30&94.63&94.85&90.38&94.63&94.63&95.08\\
Class5&81.78&81.64&84.22&86.80&\bf94.55&91.68&88.52&93.54&88.95&90.10&90.24&91.25\\
Class6&95.90&95.67&95.67&97.95&97.95&98.63&98.41&98.41&94.99&99.32&96.58&\bf99.77\\
Class7&66.56&67.32&77.89&58.06&70.81&74.95&73.20&75.16&73.09&73.31&81.37&\bf97.39\\
Class8&55.21&54.18&55.29&42.97&52.94&57.49&54.43&55.21&57.78&63.52&76.51&\bf87.80\\
Class9&53.01&52.30&54.96&71.45&79.61&82.27&68.44&78.01&72.34&81.56&84.40&\bf93.26\\
Class10&98.15&98.15&98.15&\bf99.38&\bf99.38&\bf99.38&\bf99.38&\bf99.38&96.30&\bf99.38&\bf99.38&\bf99.38\\
Class11&82.88&82.40&86.90&85.53&89.79&88.18&87.94&89.87&86.58&89.31&93.41&\bf98.63\\
Class12&50.91&51.21&61.21&77.88&83.03&82.73&81.21&81.52&73.03&82.12&79.09&\bf96.06\\
Class13&97.78&97.78&97.78&97.78&97.78&\bf100.00&97.78&97.78&97.78&95.56&\bf100.00&\bf100.00\\
Class14&79.49&79.49&87.18&74.36&92.31&82.05&82.05&94.87&58.97&84.62&\bf97.44&87.18\\
Class15&81.82&81.82&90.91&\bf100.00&\bf100.00&\bf100.00&\bf100.00&90.91&72.73&\bf100.00&90.91&\bf100.00\\
Class16&\bf100.00&\bf100.00&\bf100.00&40.00&\bf100.00&60.00&\bf100.00&\bf100.00&\bf100.00&80.00&\bf100.00&\bf100.00\\
\bottomrule[1.5pt]
\end{tabular}}
\label{tab:Indian}
\end{table*}

\subsection{Results Analysis and Discussion}
\subsubsection{Indian Pines Dataset}
Table \ref{tab:Indian} presents the classification performances of the different methods with the optimal parameter setting tuned by cross-validation on the training set using the NN classifier. Correspondingly, the classification maps are given in Fig. \ref{fig:CM_Indian} for visual assessment.

Overall, PCA provides similar performances with the baseline (OSF), as the PCA more focuses on maximizing the information but could fail to excavate the potential data structure that lies in reality. By smoothing the spatial structure of HSI, OTVCA enables better identification of the materials than OSF and PCA. For the supervised HDR methods, the classification performances of classic LDA are even lower than those previously mentioned, due to the limited amount of training samples. Holding a more powerful intra-class homogeneity and inter-class separation, LFDA obtains more competitive results by locally focusing on discriminative information, which is obviously better than those of the baseline, PCA, and LDA around 8\%. However, the features learned by LFDA are relatively difficult to be generalized, due to the small-size labeled samples. Comparatively, SELD learns a robust low-dimensional feature representation with a higher generalization ability, since unlabeled samples are involved in the process of model training. In SELD, the unlabeled information is embedded by computing the similarities between samples, which is more effective than that using the pseudo-labels (e.g., SLLDA and SSLFDA) to some extent. However, these semi-supervised methods are still bad at handling noisy data. A direct proof can be shown in Fig. \ref{fig:CM_Indian} that there exist obvious salt-and-pepper-like noises in classification maps of SELD, SLLDA, and SSLFDA. Likewise, although the SAE holds a strong nonlinear learning ability in data representation, its performance is still limited by complex spectral variability and pixel-wise feature embedding. Thanks to the spatial information modeling, CAE locally extracts the spatial information and thus obtains a relatively smooth classification result. With the benefit of a multi-linear regression system, the J-Play algorithm performs much better (at least 7\% OAs) than DNN-based nonlinear HDR (SAE and CAE). Such a strategy makes the learned features more robust against various spectral deformation and degradation, in spite of without accounting for the spatial information.

The performances of the proposed JPSA are superior to the other methods, which indicates that JPSA can learn a more discriminative and robust spectral embedding. The alignment-based SSM embedding enables us to identify the materials at a more accurate level on a small-scale training set. As shown in Fig. \ref{fig:CM_Indian}, the classification map obtained by JPSA is smoother than others, demonstrating that our method is capable of effectively aggregating the spatially contextual information in the process of HDR by means of superpixels. It is worth noting that the JPSA not only outperforms others from the whole perspective, but also obtains highly competitive results for each class, particularly for \textit{Corn}, \textit{Soybean-Notill}, \textit{Soybean-Mintill}, \textit{Soybean-Clean}, and \textit{Building-Grass-Trees} that have a dramatic improvement of about 10\% in classification accuracy.

\subsubsection{The University of Houston Dataset}
Fig. \ref{fig:CM_Houston} shows a visual comparison among the different algorithms, and the specific classification accuracies for various compared methods, which were optimally parameterized by a cross-validation as listed in Table \ref{tab:Houston}.

Generally, there is a basically consistent trend in classification performance between OSF and PCA: around 72\% OA as a baseline. For another unsupervised HDR method, OTVCA approximately yields a 2\% improvement on the basis of OSF and PCA. Owing to the use of total variation operator in OTVCA (see the smooth classification map in Fig. \ref{fig:CM_Houston}), it shares similar performances with discriminant analysis-based approaches such as LDA and LFDA. This reason why the unsupervised OTVCA is comparable to the supervised HDR methods could be, to some extent, two-fold. On one hand, local smoothing strategy is a good fit for HS feature extraction and HDR tasks; on the other hand, the small-size training set hinders the supervised LDA and LFDA finding a generalized or transferable discriminative subspace. Nevertheless, LFDA is capable of steadily performing better than OTVCA owing to the consideration of local manifold structure. This might be seen as indirect evidence to show the effectiveness of the manifold embedding in HDR. More intuitively, the performance of semi-supervised methods is superior to that of those only considering the labeled samples, where the SSLFDA achieves the best classification results. This demonstrates the effectiveness of embedding unlabeled samples in improving the generalization ability of the learned model. Although these semi-supervised methods show the discriminative power between different classes, yet there is still room for improvement in spatial information modeling and model learning ability. As a member of deep learning, SAE is capable of better reducing the gap between the original data and compressed data, thus yielding better classification performance. Another DL-based technique for HDR is CAE, which can extract a low-dimensional spectral representation with the attention of spatial contextual information. As a result, CAE performs better than the pixel-wise SAE with an about 1\% slight increase of OA. Due to the lack of modeling spectral variability, SAE or CAE fails to transfer the trained model to out-of-sample (i.e. test set) effectively, even though there is a powerful learning ability in SAE and CAE. Unlike them, J-Play adopts a multi-linear modeling strategy with the SE constraint in order to remove the spectral variabilities effectively and maintain the learned features as discriminative as possible, which results in basically the same results with CAE and slightly higher than SAE.

\begin{table*}[!t]
\centering
\caption{Quantitative performance comparisons of different algorithms on the University of Houston dataset with the optimal parameter combination in terms of OA, AA, and $\kappa$, as well as the accuracy for each class. The best one is shown in bold. JPlay$_{3}$ means a three-layered J-Play model ($m=3$), while JPSA$_{3}$ denotes a three-layered JPSA model ($m=3$).}
\resizebox{0.85\textwidth}{!}{
\begin{tabular}{p{30pt}<{\centering}||p{30pt}<{\centering}|p{30pt}<{\centering}|p{32pt}<{\centering}||p{30pt}<{\centering}|p{30pt}<{\centering}||p{30pt}<{\centering}|p{30pt}<{\centering}|p{35pt}<{\centering}||p{30pt}<{\centering}|p{30pt}<{\centering}||p{30pt}<{\centering}|p{30pt}<{\centering}}
%\begin{tabular}{c||ccc|cc|ccc|cc|cc}
\toprule[1.5pt]
Method & OSF & PCA & OTVCA & LDA & LFDA & SELD & SLLDA & SSLFDA & SAE & CAE & JPlay$_{3}$ & JPSA$_{3}$\\
\hline \hline
$d$ & $144$ & $20$ & $15$ & $14$ & $14$ & $14$ & $14$ & $14$ & $30$ & $30$ & $30$ & $30$\\
$k$ & $-$ & $-$ & $-$ & $-$ & $20$ & $20$ & $-$ & $30$ & $-$ & $-$ & $10$ & $10$ \\
$\sigma$ & $-$ & $-$ & $-$ & $-$ & $0.1$ & $0.1$ & $-$ & $0.1$ & $-$ & $-$ & $0.1$ & $0.1$ \\
$\alpha$ & $-$ & $-$ & $-$ & $-$ & $-$ & $-$ & $-$ & $-$ & $-$ & $-$ & $1$ & $1$\\
$\beta$ & $-$ & $-$ & $-$ & $-$ & $-$ & $-$ & $-$ & $-$ & $-$ & $-$ & $0.1$ & $0.1$\\
$\gamma$ & $-$ & $-$ & $-$ & $-$ & $-$ & $-$ & $-$ & $-$ & $-$ & $-$ & $0.1$ & $0.1$\\
\hline\hline
OA&72.83&72.75&74.18&74.18&75.52&77.45&77.18&78.94&79.52&80.68&80.13&\bf86.09\\
AA&76.16&76.09&77.61&79.04&79.10&80.40&79.59&82.09&82.45&83.23&82.99&\bf87.90\\
$\kappa$&0.7079&0.7071&0.7228&0.7374&0.7355&0.7555&0.7537&0.7716&0.7789&0.7905&0.7845&\bf0.8490\\
\hline \hline
Class1&82.15&82.15&82.24&81.67&81.96&81.29&81.96&82.43&82.53&82.15&\bf82.72&81.10\\
Class2&81.86&81.86&82.05&82.14&82.99&83.46&83.36&82.14&83.27&83.74&82.61&\bf84.68\\
Class3&99.60&99.60&99.60&\bf100.00&\bf100.00&\bf100.00&\bf100.00&\bf100.00&99.80&99.80&\bf100.00&99.41\\
Class4&91.76&91.76&91.86&90.44&91.00&93.75&91.57&92.42&88.83&91.48&\bf96.78&94.89\\
Class5&97.06&97.06&97.73&97.25&97.82&99.34&98.01&99.05&98.11&\bf99.91&99.62&99.72\\
Class6&95.10&95.10&94.41&99.30&99.30&\bf100.00&94.41&99.30&95.10&97.90&96.50&96.50\\
Class7&73.60&73.60&79.10&72.57&81.72&81.34&72.76&\bf85.73&79.10&78.64&80.50&78.17\\
Class8&36.37&36.37&36.56&59.92&40.65&42.17&50.71&42.55&43.59&64.20&52.23&\bf76.35\\
Class9&66.19&66.29&67.14&57.60&74.13&72.33&63.74&73.84&74.41&74.22&77.15&\bf79.70\\
Class10&49.23&49.03&48.26&57.53&42.95&45.08&57.92&49.52&58.30&51.25&53.28&\bf81.95\\
Class11&67.74&67.55&70.68&74.76&74.19&82.64&81.02&81.50&79.22&\bf85.77&75.81&74.00\\
Class12&54.27&53.60&57.64&58.79&59.46&69.36&69.55&75.60&84.53&75.70&77.52&\bf97.31\\
Class13&51.93&51.93&58.95&56.49&62.81&58.60&51.93&69.82&71.58&68.77&72.63&\bf80.00\\
Class14&97.57&97.57&99.19&99.19&99.19&99.19&99.19&99.19&\bf100.00&\bf100.00&98.79&\bf100.00\\
Class15&97.89&97.89&\bf98.73&97.89&98.31&97.46&97.67&98.31&98.31&94.93&\bf98.73&94.71\\
\bottomrule[1.5pt]
\end{tabular}}
\label{tab:Houston}
\end{table*}

\begin{table}[!t]
\centering
\caption{Classification performance (OA, AA, and $\kappa$) with the different number of learnt projections ($m$) on the two datasets.}
\resizebox{0.45\textwidth}{!}{
\begin{tabular}{p{30pt}<{\centering}|p{30pt}<{\centering}|p{30pt}<{\centering}|p{30pt}<{\centering}|p{30pt}<{\centering}|p{30pt}<{\centering}|p{30pt}<{\centering}}
%\begin{tabular}{c||ccc|cc|ccc|cc|cc}
\toprule[1.5pt]
\multirow{2}{*}{Method} & \multicolumn{3}{c|}{Indian Pines} & \multicolumn{3}{c}{University of Houston}\\
\cline{2-7} & OA & AA & $\kappa$ & OA & AA & $\kappa$\\
\hline\hline
JPSA$_{1}$&87.41&93.13&0.8565&81.75&83.82&0.8019\\
JPSA$_{2}$&90.74&94.58&0.8942&82.27&84.35&0.8074\\
JPSA$_{3}$&92.28&94.97&0.9116&\bf86.09&\bf87.90&\bf0.8490\\
JPSA$_{4}$&\bf92.98&\bf95.40&\bf0.9197&84.82&86.88&0.8353\\
JPSA$_{5}$&91.35&95.02&0.9012&84.20&86.10&0.8285\\
JPSA$_{6}$&92.76&95.23&0.9173&82.89&85.19&0.8143\\
JPSA$_{7}$&89.74&94.21&0.8831&82.44&84.56&0.8094\\
JPSA$_{8}$&90.79&94.43&0.8948&81.54&82.97&0.7997\\
\bottomrule[1.5pt]
\end{tabular}}
\label{tab:layers}
\end{table}

JPSA outperforms other HDR algorithms significantly, which indicates that the proposed method is capable of effectively approximating an optimal mapping from the original space to the label space by fully considering a trade-off between spectral discrimination and subspace robustness, thus providing a robust and discriminative low-dimensional feature representation. Further, the embedding of spatial-spectral information enables semantically meaningful object-based HS classification results. Notably, JPSA is able to more effectively eliminate the effects of shadow covered by clouds in image acquisition, compared to other methods as shown in Fig. \ref{fig:CM_Houston}. Accordingly, JPSA also shows the superiority in identifying different materials, as quantified in Table \ref{tab:Houston}, especially for those challenging classes, such as \textit{Commercial}, \textit{Highway}, and \textit{Parking Lot1}.

\subsection{Parameter Sensitivity Analysis of JPSA}
The quality of low-dimensional feature embedding, to some extent, depends on the parameter selection, it is, therefore, indispensable to investigate the sensitivity of parameter setting in JPSA. Five main parameters involved in the JPSA, which need to be emphatically analyzed and discussed, would result in a significant effect on dimension-reduced features and even final classification results. They include three regularization parameters ($\alpha$, $\beta$, and $\gamma$) in Eq. (\ref{eq3}), subspace dimension ($d$), and the number of layers ($m$).
\begin{figure*}[!t]
	  \centering
		\subfigure{
			\includegraphics[width=0.99\textwidth]{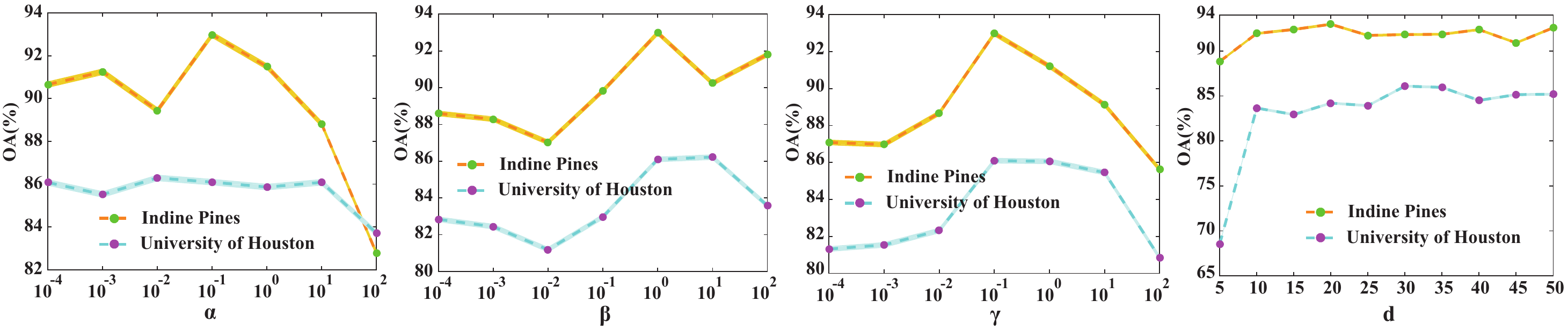}
		}
        \caption{Parameter sensitivity analysis of JPSA for three regularization parameters ($\alpha$, $\beta$, and $\gamma$) and the subspace dimension ($d$) on the two datasets.}
\label{fig:Param}
\end{figure*}

Significantly, we start to analyze the effects of different $m$ for JPSA. With the different number of learnt projections, we successively specify the proposed model as JPSA$_{1}$, \dots, JPSA$_{l}$, \dots, JPSA$_{m}$, $\forall l=1,...,m$. To investigate the trend of OAs, $m$ is uniformly set up to $8$ on the two datasets. We experimentally set the number of clusters in SLIC as 10$\%$ of the total samples. As listed in Table \ref{tab:layers}, with the increase of $m$, the performances of JPSA with SSM embedding steadily increase to the best with around $3$ layers for both datasets and then gradually decrease with a slight perturbation. This might be explained by over-fitting and error accumulation of the model in the multi-layered regression process, since our model is only trained on a limited number of samples. Note that more results about the JPlay in terms of the parameter $m$ can be found in \cite{hong2018joint}, and the code is openly available from the link: \url{https://github.com/danfenghong/ECCV2018_J-Play}.

Apart from the parameter $m$, the regularization parameters and subspace dimension also play a crucial role in improving the model's performance. More specifically, the resulting quantitative analysis of the two datasets is given in Fig. \ref{fig:Param}, where the parameter combinations of $(\alpha=1, \; \beta=0.1, \; \gamma=0.1, \; d=20)$ and $(\alpha=1, \; \beta=0.1, \; \gamma=0.1, \; d=30)$ achieve the best classification performance on the test sets for the first and second datasets, respectively. The resulting parameter selection for the two sets of datasets is basically consistent with that determined by 10-fold cross-validation on the training set (please see Section III.B for more details). The cross-validation is, therefore, an effective strategy to automatically determine the model's parameters so that other researchers are able to produce the results for their own tasks. More specifically, the optimal parameters can be determined by testing all of the parameter combinations. Furthermore, we only show the two-dimensional figures (see Fig. \ref{fig:Param}) for the convenience of visualization, where other variables are set to be the optimal ones except for the current investigated variable.

Moreover, we can observe from Fig. \ref{fig:Param}, that with the increase of $d$, the JPSA's performance increases to the optimal value with the dimension of 20 for the Indian Pines dataset and 30 for the University of Houston dataset, respectively, then starts to reach a relatively stable state, and finally decreases with a slight perturbation when the subspace dimension is approaching to that of original spectral signature. For the variable $\alpha$ that mainly controls the prediction errors between the input data and labels, it is a very important factor that needs to be carefully considered in the model learning, since the setting of $\alpha$ is sensitive to the feature embedding and even to the final classification results. Similarly, the terms of SR and SSM alignment also have great effects on the classification performance, which indicates the importance of the two terms. What's more, the subspace dimension is a noteworthy factor as well, although the OAs with different dimensions are relatively stable when the variable $d$ reaches a larger value (e.g., 10).
\begin{table}[!t]
\centering
\caption{Ablation analysis of JPSA with a progressive combination of different terms on the two datasets.}
\resizebox{0.45\textwidth}{!}{
\begin{tabular}{p{35pt}<{\centering}|p{30pt}<{\centering}|p{30pt}<{\centering}|p{30pt}<{\centering}|p{30pt}<{\centering}|p{30pt}<{\centering}|p{30pt}<{\centering}}
%\begin{tabular}{c||ccc|cc|ccc|cc|cc}
\toprule[1.5pt]
\multirow{2}{*}{Terms} & \multicolumn{3}{c|}{Indian Pines} & \multicolumn{3}{c}{University of Houston}\\
\cline{2-7} & OA & AA & $\kappa$ & OA & AA & $\kappa$\\
\hline\hline
None &87.92&93.16&0.8623&82.22&84.74&0.8068\\
SR &89.40&93.81&0.8791&84.44&86.60&0.8310\\
SR+SSM&\bf92.98&\bf95.40&\bf0.9197&\bf86.09&\bf87.90&\bf0.8490\\
\bottomrule[1.5pt]
\end{tabular}}
\label{tab:Ablation}
\end{table}
\subsection{Ablation Studies of JPSA}
Additionally, we analyze the performance gain of JPSA by step-wise adding the different components, i.e., SR term, SSM alignment term, etc. Table \ref{tab:Ablation} details the increasing performance when different terms are fused. As it turns out successively embedding each component into the JPSA would lead to a progressive enhancement in feature representation ability. This demonstrates the advancement and effectiveness of the proposed JPSA model for HDR.

\section{Conclusion}
In this paper, we proposed a joint and progressive subspace analysis (JPSA) technique to learn an optimal mapping for effective HS data compression along the spectral dimension. JPSA is expected to find a discriminative subspace where the samples can be semantically (label information) and structurally (SSM or topology perseveration and alignment) represented and thereby be better classified. Oriented by assessing pixel-wise HS classification performances, we conduct extensive experiments using JPSA in comparison with some previous state-of-the-art HDR methods. The desirable results using JPSA demonstrate its superiority and effectiveness, particularly in handling various complex spectral variabilities compared to other nonlinear DR techniques (e.g., DL-based methods). In the future, we will further develop and apply the JPSA framework to the multi-modality learning.
\appendices
\section{Solution to AutoRULe}
The solution to problem (\ref{eq12}) can be transferred to equivalently solve the problem (\ref{eq13}) with ADMM. Considering the fact that the object function in Eq. (\ref{eq13}) is not convex with respect to all variables simultaneously, but it is a convex problem regarding the separate variable when other variables are fixed, therefore we successively minimize $\mathscr{L}_{\mu}$ (Eq. (\ref{eq13})) with respect to $\mathbf{\Theta}_{l},\mathbf{H},\mathbf{G},\mathbf{Q}, \mathbf{S}, \{\mathbf{\Lambda}_{n}\}_{n=1}^{4}$ as follows:

{\it $\mathbf{\Theta}_{l}$ problem:} The optimization problem for $\mathbf{\Theta}$ is
\begin{equation}
\label{eq21}
\begin{aligned}
&\mathop{\min}_{\mathbf{\Theta}_{l}}\frac{\eta}{2}\tr(\mathbf{\Theta}_{l}\mathbf{\tilde{X}}_{l-1}\mathbf{L}^{f}\mathbf{\tilde{X}}_{l-1}^{\T}\mathbf{\Theta}_{l}^{\T})\!+\!\frac{\mu}{2}\norm{\mathbf{H}-\mathbf{\Theta}_{l}\mathbf{\tilde{X}}_{l-1}}_{\F}^{2}\\
&\!+\!\mathbf{\Lambda}_{1}^{\T}(\mathbf{H}-\mathbf{\Theta}_{l}\mathbf{\tilde{X}}_{l-1})\!+\!\frac{\mu}{2}\norm{\mathbf{G}-\mathbf{\Theta}_{l}}_{\F}^{2}\!+\!\mathbf{\Lambda}_{2}^{\T}(\mathbf{G}-\mathbf{\Theta}_{l})\\
&\!+\!\frac{\mu}{2}\norm{\mathbf{Q}-\mathbf{\Theta}_{l}\mathbf{\tilde{X}}_{l-1}}_{\F}^{2}\!+\!\mathbf{\Lambda}_{3}^{\T}(\mathbf{Q}-\mathbf{\Theta}_{l}\mathbf{\tilde{X}}_{l-1})\!+\!l_{R}^{+}(\mathbf{Q})\\
&\!+\!\frac{\mu}{2}\norm{\mathbf{S}-\mathbf{\Theta}_{l}\mathbf{\tilde{X}}_{l-1}}_{\F}^{2}\!+\!\mathbf{\Lambda}_{4}^{\T}(\mathbf{S}-\mathbf{\Theta}_{l}\mathbf{\tilde{X}}_{l-1})\!+\!l_{R}^{\sim}(\mathbf{S}),
\end{aligned}
\end{equation}
which has a closed-form solution:
\begin{equation}
\label{eq22}
\mathbf{\Theta}_{l}\leftarrow
\begin{aligned}
&\left(
\begin{aligned}
       &\mu\mathbf{H}\mathbf{\tilde{X}}_{l-1}^{\T}+\mu\mathbf{G}+\mu\mathbf{Q}\mathbf{\tilde{X}}_{l-1}^{\T}+\mu\mathbf{P}\mathbf{\tilde{X}}_{l-1}^{\T}\\
       &+\mathbf{\Lambda}_{1}\mathbf{\tilde{X}}_{l-1}^{\T}+\mathbf{\Lambda}_{2}+\mathbf{\Lambda}_{3}\mathbf{\tilde{X}}_{l-1}^{\T}+\mathbf{\Lambda}_{4}\mathbf{\tilde{X}}_{l-1}^{\T}\\
       \end{aligned}
       \right)\\
       &\times(\eta(\mathbf{\tilde{X}}_{l-1}\mathbf{L}^{f}\mathbf{\tilde{X}}_{l-1}^{\T})+3\mu(\mathbf{\tilde{X}}_{l-1}\mathbf{\tilde{X}}_{l-1}^{\T})+\mu\mathbf{I})^{-1}.
\end{aligned}
\end{equation}

{\it $\mathbf{H}$ problem:} The variable $\mathbf{H}$ can be estimated by solving the following problem:
\begin{equation}
\label{eq23}
\begin{aligned}
\mathop{\min}_{\mathbf{H}}\frac{1}{2}\norm{\mathbf{\tilde{X}}_{l-1}&-\mathbf{G}^{\T}\mathbf{H}}_{\F}^{2}+\frac{\mu}{2}\norm{\mathbf{H}-\mathbf{\Theta}_{l}\mathbf{\tilde{X}}_{l-1}}_{\F}^{2}\\
&+\mathbf{\Lambda}_{1}^{\T}(\mathbf{H}-\mathbf{\Theta}_{l}\mathbf{\tilde{X}}_{l-1}),
\end{aligned}
\end{equation}
its analytical solution is given by
\begin{equation}
\label{eq24}
\begin{aligned}
       \mathbf{H}\leftarrow(\mathbf{G}\mathbf{G}^{\T}+\mu\mathbf{I})^{-1}(\mathbf{G}\mathbf{\tilde{X}}_{l-1}+\mu\mathbf{\Theta}_{l}\mathbf{\tilde{X}}_{l-1}-\mathbf{\Lambda}_{1}).
\end{aligned}
\end{equation}

{\it $\mathbf{G}$ problem:} The optimization problem can be written as
\begin{equation}
\label{eq25}
\begin{aligned}
\mathop{\min}_{\mathbf{G}}\frac{\mu}{2}\norm{\mathbf{G}-\mathbf{\Theta}_{l}}_{\F}^{2}+\mathbf{\Lambda}_{2}^{\T}(\mathbf{G}-\mathbf{\Theta}_{l}),
\end{aligned}
\end{equation}
which can be effectively solved as
\begin{equation}
\label{eq26}
\begin{aligned}
       \mathbf{G}\leftarrow(\mathbf{H}\mathbf{H}^{\T}+\mu\mathbf{I})^{-1}(\mathbf{H}\mathbf{\tilde{X}}_{i}+\mu\mathbf{\Theta}_{l}-\mathbf{\Lambda}_{2}).
\end{aligned}
\end{equation}

{\it $\mathbf{Q}$ problem:} The optimization problem of $\mathbf{Q}$ is
\begin{equation}
\label{eq27}
\begin{aligned}
\mathop{\min}_{\mathbf{Q}}\frac{\mu}{2}\norm{\mathbf{Q}-\mathbf{\Theta}_{l}\mathbf{\tilde{X}}_{l-1}}_{\F}^{2}+\mathbf{\Lambda}_{3}^{\T}(\mathbf{Q}-\mathbf{\Theta}_{l}\mathbf{\tilde{X}}_{l-1})+l_{R}^{+}(\mathbf{Q}).
\end{aligned}
\end{equation}
Here the update rule for $\mathbf{Q}$ can be expressed as
\begin{equation}
\label{eq28}
\begin{aligned}
       \mathbf{Q}\leftarrow \max(\mathbf{\Theta}_{l}\mathbf{\tilde{X}}_{l-1}-\mathbf{\Lambda}_{3}/\mu,\mathbf{0}).
\end{aligned}
\end{equation}

{\it $\mathbf{S}$ problem:} The variable $\mathbf{S}$ is estimated by solving
\begin{equation}
\label{eq29}
\begin{aligned}
\mathop{\min}_{\mathbf{S}}\frac{\mu}{2}\norm{\mathbf{S}-\mathbf{\Theta}_{l}\mathbf{\tilde{X}}_{l-1}}_{\F}^{2}+\mathbf{\Lambda}_{4}^{\T}(\mathbf{S}-\mathbf{\Theta}_{l}\mathbf{\tilde{X}}_{l-1})+l_{R}^{\sim}(\mathbf{S}),
\end{aligned}
\end{equation}
whose solution can be updated in each iteration by the vector-based projection operator of Eq. (\ref{eq15}):
\begin{equation}
\label{eq30}
\begin{aligned}
       \mathbf{S}\leftarrow prox_{f}(\mathbf{\Theta}_{l}\mathbf{\tilde{X}}_{l-1}-\mathbf{\Lambda}_{4}/\mu).
\end{aligned}
\end{equation}

{\it Lagrange multipliers ($\{\mathbf{\Lambda}_{i}\}_{i=1}^{4}$) update:} Before stepping into the next iteration, Lagrange multipliers are updated by
\begin{equation}
\label{eq31}
\begin{aligned}
        &\mathbf{\Lambda}_{1}=\mathbf{\Lambda}_{1}+\mu(\mathbf{H}-\mathbf{\Theta}_{i}\mathbf{\tilde{X}}_{l-1}),\; \mathbf{\Lambda}_{2}=\mathbf{\Lambda}_{2}+\mu(\mathbf{G}-\mathbf{\Theta}_{i}),\\
        &\mathbf{\Lambda}_{3}=\mathbf{\Lambda}_{3}+\mu(\mathbf{Q}-\mathbf{\Theta}_{i}\mathbf{\tilde{X}}_{l-1}),\;
        \mathbf{\Lambda}_{4}=\mathbf{\Lambda}_{4}+\mu(\mathbf{P}-\mathbf{\Theta}_{i}\mathbf{\tilde{X}}_{l-1}).
\end{aligned}
\end{equation}

\section*{Acknowledgments}

The authors would like to thank the Hyperspectral Image Analysis group and the NSF Funded Center for Airborne Laser Mapping (NCALM) at the University of Houston for providing the CASI University of Houston dataset. The authors would like to express their appreciation to Prof. D. Cai and Dr. C. Wang for providing MATLAB codes for LPP and manifold alignment algorithms.

\bibliographystyle{ieeetr}
\bibliography{HDF_ref}

\begin{IEEEbiography}[{\includegraphics[width=1in,height=1.25in,clip,keepaspectratio]{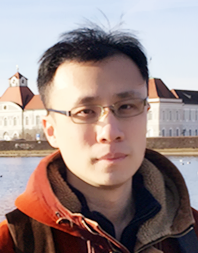}}]{Danfeng Hong}
(S'16--M'19) received the M.Sc. degree (summa cum laude) in computer vision, College of Information Engineering, Qingdao University, Qingdao, China, in 2015, the Dr. -Ing degree (summa cum laude) in Signal Processing in Earth Observation (SiPEO), Technical University of Munich (TUM), Munich, Germany, in 2019.

Since 2015, he has been a Research Associate with Remote Sensing Technology Institute (IMF), German Aerospace Center (DLR), wessling, Germany. He is a Research Scientist and leads a Spectral Vision Working Group at IMF, DLR, and also an Adjunct Scientist at the GIPSA-lab, Grenoble INP, CNRS, Univ. Grenoble Alpes, Grenoble, France.

His research interests include signal / image processing and analysis, hyperspectral remote sensing, machine / deep learning, artificial intelligence and their applications in Earth Vision.
\end{IEEEbiography}

\vskip -2\baselineskip plus -1fil

\begin{IEEEbiography}[{\includegraphics[width=1in,height=1.25in,clip,keepaspectratio]{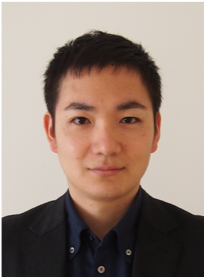}}]{Naoto Yokoya} (S'10--M'13) received the M.Eng. and Ph.D. degrees from the Department of Aeronautics and Astronautics, the University of Tokyo, Tokyo, Japan, in 2010 and 2013, respectively.

He is currently a Lecturer at the University of Tokyo and a Unit Leader at the RIKEN Center for Advanced Intelligence Project, Tokyo, Japan, where he leads the Geoinformatics Unit. He was an Assistant Professor at the University of Tokyo from 2013 to 2017. In 2015-2017, he was an Alexander von Humboldt Fellow, working at the German Aerospace Center (DLR), Oberpfaffenhofen, and Technical University of Munich (TUM), Munich, Germany. His research is focused on the development of image processing, data fusion, and machine learning algorithms for understanding remote sensing images, with applications to disaster management.

Dr. Yokoya won the first place in the 2017 IEEE Geoscience and Remote Sensing Society (GRSS) Data Fusion Contest organized by the Image Analysis and Data Fusion Technical Committee (IADF TC). He is the Chair (2019-2021) and was a Co-Chair (2017-2019) of IEEE GRSS IADF TC and also the secretary of the IEEE GRSS All Japan Joint Chapter since 2018. He is an Associate Editor for the IEEE Journal of Selected Topics in Applied Earth Observations and Remote Sensing (JSTARS) since 2018. He is/was a Guest Editor for the IEEE JSTARS in 2015-2016, for Remote Sensing in 2016-2020, and for the IEEE Geoscience and Remote Sensing Letters (GRSL) in 2018-2019.
\end{IEEEbiography}

\vskip -2\baselineskip plus -1fil

\begin{IEEEbiography}[{\includegraphics[width=1in,height=1.25in,clip,keepaspectratio]{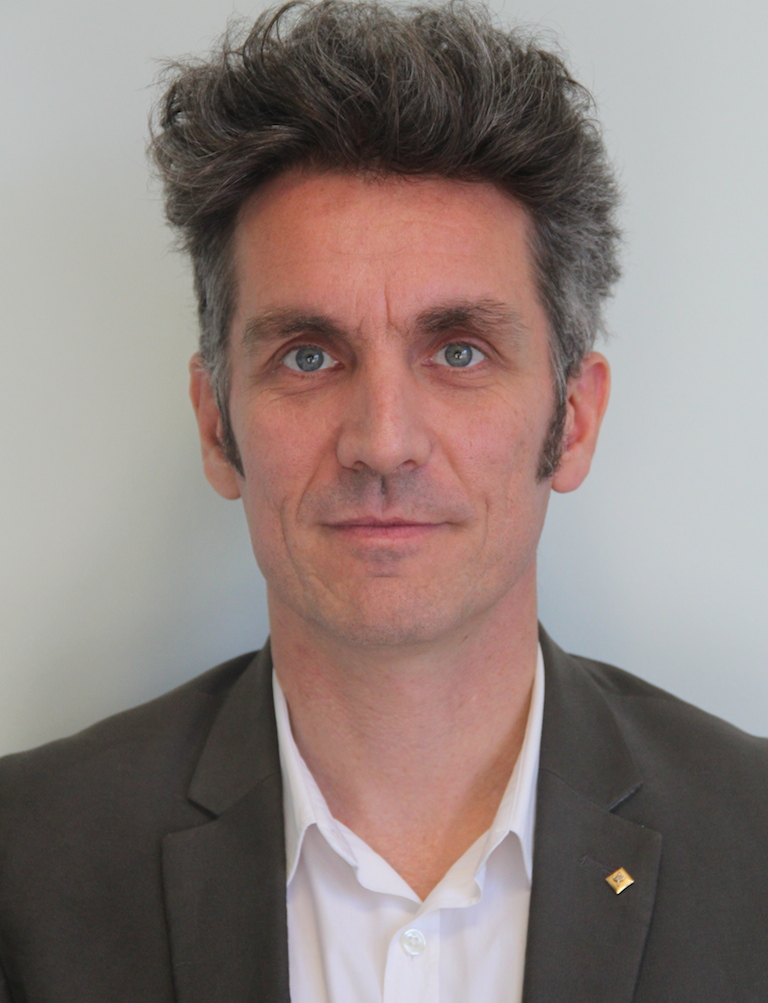}}]{Jocelyn Chanussot}
(M'04--SM'04--F'12) received the M.Sc. degree in electrical engineering from the Grenoble Institute of Technology (Grenoble INP), Grenoble, France, in 1995, and the Ph.D. degree from the Université de Savoie, Annecy, France, in 1998. Since 1999, he has been with Grenoble INP, where he is currently a Professor of signal and image processing. His research interests include image analysis, hyperspectral remote sensing, data fusion, machine learning and artificial intelligence. He has been a visiting scholar at Stanford University (USA), KTH (Sweden) and NUS (Singapore). Since 2013, he is an Adjunct Professor of the University of Iceland. In 2015-2017, he was a visiting professor at the University of California, Los Angeles (UCLA). He holds the AXA chair in remote sensing and is an Adjunct professor at the Chinese Academy of Sciences, Aerospace Information research Institute, Beijing.

Dr. Chanussot is the founding President of IEEE Geoscience and Remote Sensing French chapter (2007-2010) which received the 2010 IEEE GRS-S Chapter Excellence Award. He has received multiple outstanding paper awards. He was the Vice-President of the IEEE Geoscience and Remote Sensing Society, in charge of meetings and symposia (2017-2019). He was the General Chair of the first IEEE GRSS Workshop on Hyperspectral Image and Signal Processing, Evolution in Remote sensing (WHISPERS). He was the Chair (2009-2011) and  Cochair of the GRS Data Fusion Technical Committee (2005-2008). He was a member of the Machine Learning for Signal Processing Technical Committee of the IEEE Signal Processing Society (2006-2008) and the Program Chair of the IEEE International Workshop on Machine Learning for Signal Processing (2009). He is an Associate Editor for the IEEE Transactions on Geoscience and Remote Sensing, the IEEE Transactions on Image Processing and the Proceedings of the IEEE. He was the Editor-in-Chief of the IEEE Journal of Selected Topics in Applied Earth Observations and Remote Sensing (2011-2015). In 2014 he served as a Guest Editor for the IEEE Signal Processing Magazine. He is a Fellow of the IEEE, a member of the Institut Universitaire de France (2012-2017) and a Highly Cited Researcher (Clarivate Analytics/Thomson Reuters, 2018-2019).

\end{IEEEbiography}

\vskip -2\baselineskip plus -1fil

\begin{IEEEbiography}[{\includegraphics[width=1in,height=1.25in,clip,keepaspectratio]{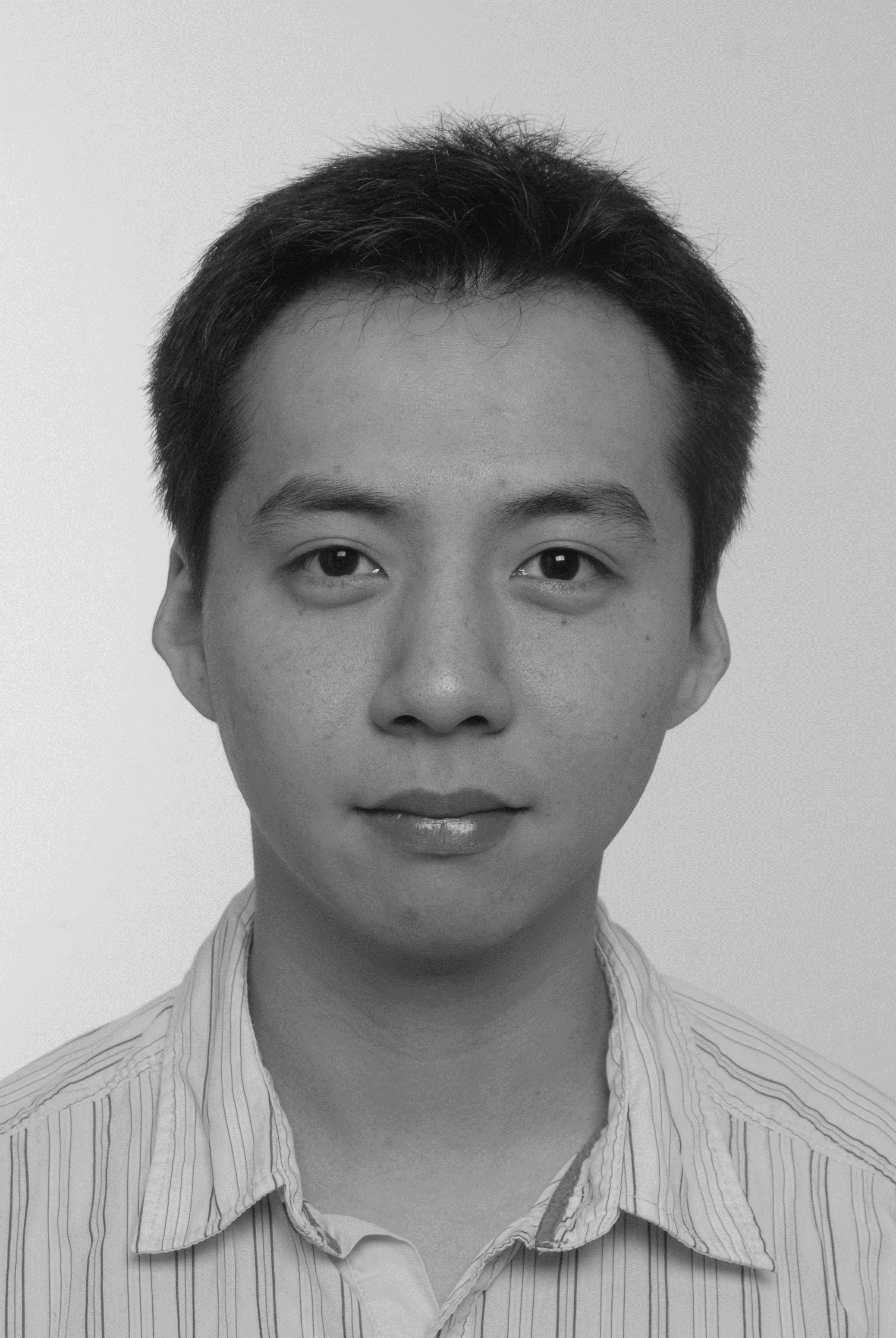}}]{Jian Xu}(M'16) received the B.E. degree in geographic information systems from Hohai University, Nanjing, China, in 2004, and the M.S. degree in Earth-oriented space science and technology and the Ph.D. degree in atmospheric remote sensing from Technical University of Munich (TUM), Munich, Germany, in 2009 and 2015, respectively.

Since 2010, he has been with the Remote Sensing Technology Institute (IMF), German Aerospace Center (DLR), Oberpfaffenhofen, Germany. His research interests include remote sensing of atmospheric temperature and trace gases, radiative transfer modeling, and ill-posed inverse problems.
\end{IEEEbiography}

\vskip -2\baselineskip plus -1fil

\begin{IEEEbiography}[{\includegraphics[width=1in,height=1.25in,clip,keepaspectratio]{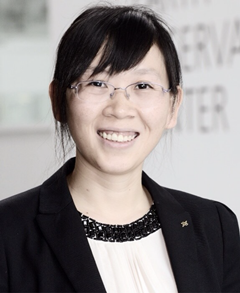}}]{Xiao Xiang Zhu}(S'10--M'12--SM'14) received the Master (M.Sc.) degree, her doctor of engineering (Dr.-Ing.) degree and her ``Habilitation'' in the field of signal processing from Technical University of Munich (TUM), Munich, Germany, in 2008, 2011 and 2013, respectively.

She is currently the Professor for Signal Processing in Earth Observation (www.sipeo.bgu.tum.de) at Technical University of Munich (TUM) and German Aerospace Center (DLR); the head of the department ``EO Data Science'' at DLR's Earth Observation Center; and the head of the Helmholtz Young Investigator Group ``SiPEO'' at DLR and TUM. Prof. Zhu was a guest scientist or visiting professor at the Italian National Research Council (CNR-IREA), Naples, Italy, Fudan University, Shanghai, China, the University  of Tokyo, Tokyo, Japan and University of California, Los Angeles, United States in 2009, 2014, 2015 and 2016, respectively. Her main research interests are remote sensing and Earth observation, signal processing, machine learning and data science, with a special application focus on global urban mapping.

Dr. Zhu is a member of young academy (Junge Akademie/Junges Kolleg) at the Berlin-Brandenburg Academy of Sciences and Humanities and the German National  Academy of Sciences Leopoldina and the Bavarian Academy of Sciences and Humanities. She is an associate Editor of IEEE Transactions on Geoscience and Remote Sensing.
\end{IEEEbiography}

\end{document}